\documentclass[10pt,journal,compsoc]{IEEEtran}
\usepackage{amsmath,amsfonts}
\usepackage{algorithmic}
\usepackage{algorithm}
\usepackage{array}
\usepackage[caption=false,font=normalsize,labelfont=sf,textfont=sf]{subfig}
\usepackage{textcomp}
\usepackage{stfloats}
\usepackage{url}
\usepackage{verbatim}
\usepackage{graphicx}
\usepackage{cite}
\usepackage{times}
\usepackage{epsfig}
\usepackage{amsmath}
\usepackage{amssymb}
\usepackage{booktabs}
\usepackage{bbding}
\usepackage{enumitem}
\usepackage[dvipsnames]{xcolor}
\usepackage{makecell}
\usepackage{multirow}
\usepackage{amsthm}
\usepackage{mathrsfs}
\usepackage{diagbox}
\usepackage{bm}
\usepackage{newfloat}
\usepackage{xspace}

\usepackage{caption} 
\definecolor{citecolor}{RGB}{0, 113, 188}
\usepackage[pagebackref,breaklinks,colorlinks,citecolor=citecolor]{hyperref}
\newcommand{\system}{OmniTracker\xspace}

\newcommand*{\eg}{\emph{e.g.}\@\xspace}

\newcommand*{\ie}{\emph{i.e.}\@\xspace}
\newcommand*{\etal}{\emph{et al.}\@\xspace}

\begin{document}

\title{\system: Unifying Visual Object Tracking by Tracking-with-Detection}

\author{Junke Wang$^{*}$,~Zuxuan Wu$^{*}$~\IEEEmembership{Member,~IEEE},~Dongdong Chen~\IEEEmembership{Member,~IEEE}, \\~Chong Luo~\IEEEmembership{Member,~IEEE},~Xiyang Dai, ~Lu Yuan, ~Yu-Gang Jiang$^{\dagger}$~\IEEEmembership{Fellow,~IEEE}
\\

\IEEEcompsocitemizethanks{
\IEEEcompsocthanksitem $^{*}$ Equal contributions.
\IEEEcompsocthanksitem $^{\dagger}$ Corresponding author.
\IEEEcompsocthanksitem Junke Wang, Zuxuan Wu, and Yu-Gang Jiang are with Shanghai Key Lab of Intelligent Information Processing and School of Computer Science, Fudan University. 
\\ Email: wangjk21@m.fudan.edu.cn, \{zxwu,ygj\}@fudan.edu.cn. 
\IEEEcompsocthanksitem Dongdong Chen, Xiyang Dai, and Lu Yuan are with Microsoft Research, Redmond. 
\\ Email: cddlyf@gmail.com, \{xidai,luyuan\}@microsoft.com.
\IEEEcompsocthanksitem Chong Luo is with Microsoft Research, Asia. \\ Email: cluo@microsoft.com.

}% <-this % stops an unwanted space
}

% The paper headers
\markboth{}%
{Shell \MakeLowercase{\textit{et al.}}: A Sample Article Using IEEEtran.cls for IEEE Journals}

\IEEEtitleabstractindextext{%
\begin{abstract}
Visual Object Tracking (VOT) aims to estimate the positions of target objects in a video sequence, which is an important vision task with various real-world applications. Depending on whether the initial states of target objects are specified by provided annotations in the first frame or the categories, VOT could be classified as instance tracking (\eg, SOT and VOS) and category tracking (\eg, MOT, MOTS, and VIS) tasks. Different definitions have led to divergent solutions for these two types of tasks, resulting in redundant training expenses and parameter overhead. In this paper, combing the advantages of the best practices developed in both communities, we propose a novel tracking-with-detection paradigm, where tracking supplements appearance priors for detection and detection provides tracking with candidate bounding boxes for the association. Equipped with such a design, a unified tracking model, \system, is further presented to resolve all the tracking tasks with a fully shared network architecture, model weights, and inference pipeline, eliminating the need for task-specific architectures and reducing redundancy in model parameters. We conduct extensive experimentation on seven prominent tracking datasets of different tracking tasks, including LaSOT, TrackingNet, DAVIS16-17, MOT17, MOTS20, and YTVIS19, and demonstrate that \system achieves on-par or even better results than both task-specific and unified tracking models.
\end{abstract}

}

\maketitle
\IEEEdisplaynontitleabstractindextext

\section{Introduction}
\label{sec:intro}
\IEEEPARstart{A}{s} a fundamental task in computer vision, visual object tracking (VOT)~\cite{bertinetto2016fully,li2019siamrpn++,dicle2013way,bae2014robust,bergmann2019tracking} enjoys a wide range of application prospects, such as augmented reality, autonomous driving, and interactive systems~\cite{wang2023chatvideo}. Currently, VOT can be divided into two categories: 1) instance tracking (\ie, SOT~\cite{chen2020siamese} and VOS~\cite{oh2019video}), where the target objects of arbitrary classes are specified by the annotations in the first frame, and 2) category tracking (\ie, MOT~\cite{xiang2015learning}, MOTS~\cite{voigtlaender2019mots}, and VIS~\cite{yang2019video}), where all objects of specific categories are expected to be detected in a sequence and associated between adjacent frames. The divergent setups require customized methods with carefully designed architectures and hyper-parameters for each tracking task, leading to complex training and redundant parameters~\cite{yan2022towards,wang2022efficient}. In contrast, humans possess the capability to address various tracking tasks by nature. The increasing demand for human-like AI has motivated us to consider the possibility of designing a unified model for various types of tracking tasks.

\begin{figure*}[t]
\centering
\includegraphics[width=\linewidth]{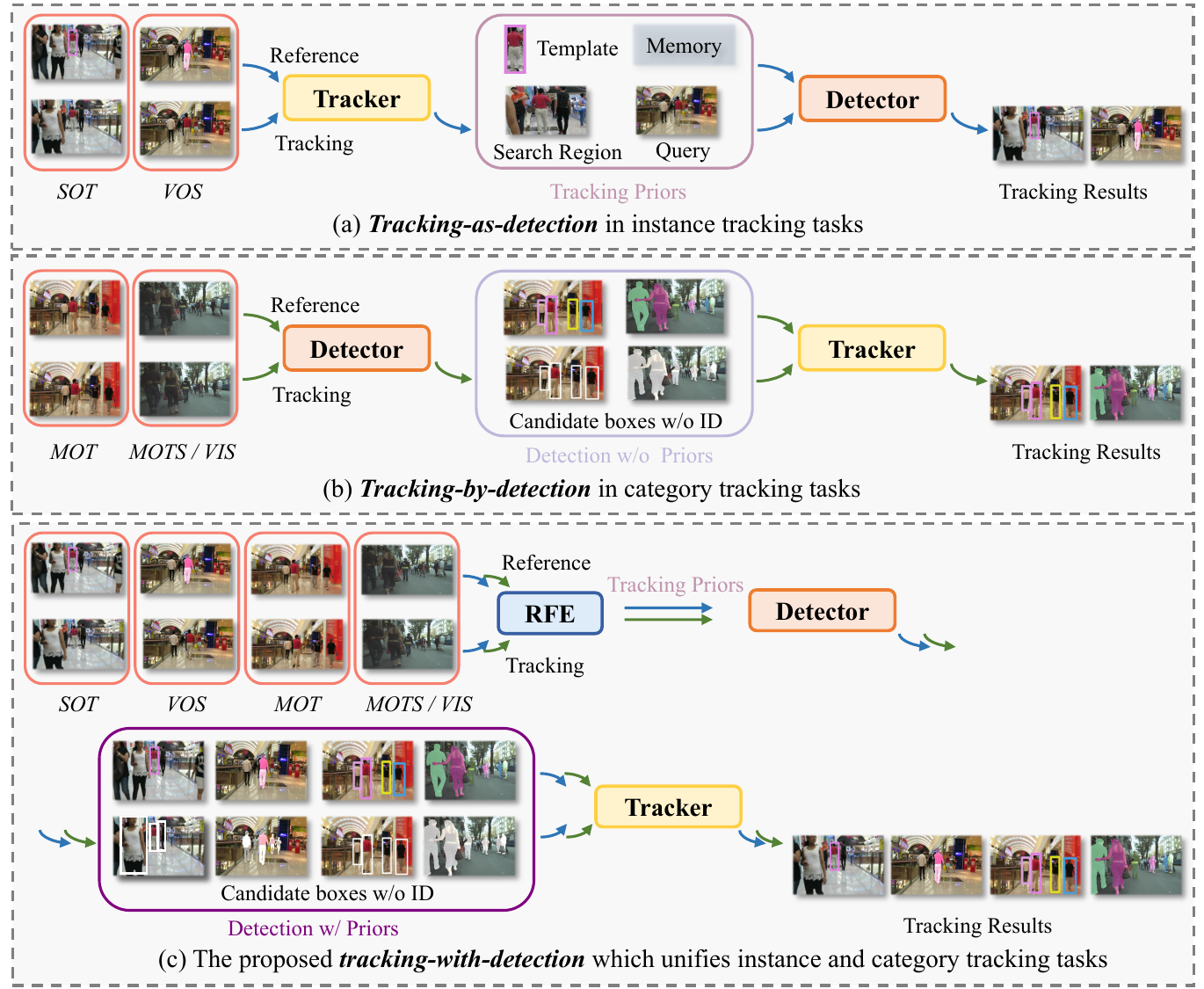}
\caption{Comparisons between different tracking paradigms. In \textit{tracking-as-detection}, the tracker delineates a search region or matches with the memory for the detector, while in \textit{tracking-by-detection}, the detector predicts the bounding boxes for the tracker to associate. We combine the advantages of them and propose a novel \textit{tracking-with-detection}, where a Reference-guided Feature Enhancement (RFE) module supplements the detector with the appearance priors, and the tracker then associates all the detected boxes with the existing trajectories according to their spatial and appearance correlation.}
\label{fig:conceptual}
\end{figure*}

Despite its promise, realizing the above goal requires non-trivial effort considering the distinction between the mainstream solutions developed in both categories of tracking tasks. Instance tracking typically treats the tracking of specified targets as a detection problem given the template bounding boxes, which we term as \textit{tracking-as-detection}. Specifically, they either crop a search region from the tracking frame based on the trajectory for SOT~\cite{li2019siamrpn++,yan2021learning,mayer2021learning,lin2021swintrack}, or match with the spatial-temporal memory which stores historical predictions for VOS~\cite{oh2019video,cheng2021stcn,cheng2022xmem,wang2022look}. The search region or memory readout is then fed into a detection head~\cite{carion2020end,ge2021yolox} or mask decoder~\cite{cheng2021stcn} to predict the bounding boxes or masks of target objects directly. Category tracking ~\cite{voigtlaender2019mots,athar2020stem,cao2020sipmask,li2021spatial,wu2022defense,zhang2022bytetrack,aharon2022bot}, on the other hand, extensively adopts the \textit{tracking-by-detection} paradigm by first sequentially detecting all objects of specific classes on each frame and then linking them based on the spatial correlation and appearance similarity. We summarize the pipelines of both paradigms in Figure~\ref{fig:conceptual} (a),(b), which illustrates that the critical difference between them lies in the role of ``detector'' and ``tracker'', \ie, whether the ``tracker'' supplements the ``detector'' with spatial and appearance priors, or the ``detector'' provides the ``tracker'' with candidate bounding boxes for the association.

In this paper, we argue that neither of these two manners fully captures the essence of tracking: unlike object detection in the image domain, tracking needs to utilize spatial and visual cues to match the detected objects with existing trajectories. The results could, in turn, provide a crucial reference for the detection in the next frame. In contrast, purely exploiting the priors from the ``tracker'' to assist the ``detector'' (tracking-as-detection) would often result in trajectory deviations after a few tracking failures~\cite{yan2022towards}, and simply applying the ``detector'' to predict the boxes on each frame independently for the ``tracker'' (tracking-by-detection) neglects the important temporal information in the detection stage, which is especially helpful for some challenging video scenarios. 

To address this issue, we introduce a new \textit{tracking-with-detection} paradigm, where a Reference-guided Feature Enhancement (RFE) module is introduced to supplement the detector with appearance priors obtained from the tracker\footnote{We abandon the location priors during training because they can be utilized during inference (\eg, via Kalman filter), but using them during training may help the network learn shortcuts and hurt the appearance representation learning. To force the network to learn stronger appearance representation, we deliberately ignore the location priors during training to enjoy the synergy of location and appearance information during inference.}. When deployed to the instance tracking tasks, we enhance the tracking frame with RoIAlign~\cite{he2017mask} features of the previously tracked boxes through cross-attention. For the category tracking tasks, considering the target objects are often occluded or blurred, we instead adopt the downsampled feature map of the previous frame. With this, the RFE module could not only adapt to different tracking tasks with the same set of parameters but also function as a task indicator for the detector. Finally, the enhanced features are input to the detector to perform object detection on the full image rather than a cropped area. Such a paradigm enables the unification of various tracking tasks by inheriting the advantages of both tracking-as-detection and tracking-by-detection.

Equipped with the proposed paradigm, we further present \system, a unified tracking model. It is built on top of the Deformable DETR~\cite{zhu2021deformable} and supports five tracking tasks, including SOT, VOS, MOT, MOTS, and VIS, within a \emph{fully shared network architecture, model weights, and inference pipeline}. Recent literature~\cite {meng2021conditional,wang2022anchor} suggests that object queries in transformer detectors can be enriched with abundant appearance and positional information through interaction with local image features. Inspired by this, we combine the well-learned queries with their corresponding RoI~\cite{he2017mask} features as the identity embedding of different instances. These embeddings are supervised using a contrastive ReID loss to learn how to associate detected objects between different frames. During inference, we maintain a memory bank for each detected instance to store historical identity embeddings for long-range matching. Extensive experiments are conducted on 7 popular tracking benchmarks, including LaSOT~\cite{fan2019lasot}, TrackingNet~\cite{muller2018trackingnet}, DAVIS16-17~\cite{pont20172017}, MOT17~\cite{milan2016mot16}, MOTS20~\cite{voigtlaender2019mots}, and YTVIS19~\cite{yang2019video}, and the results demonstrate \system achieves new state-of-the-art or at least competitive results on various tracking tasks.

%-------------------------------------------------------------------------
\section{Related Work}
\label{sec:related}

\begin{figure*}[t]
  \centering
   \includegraphics[width=\linewidth]{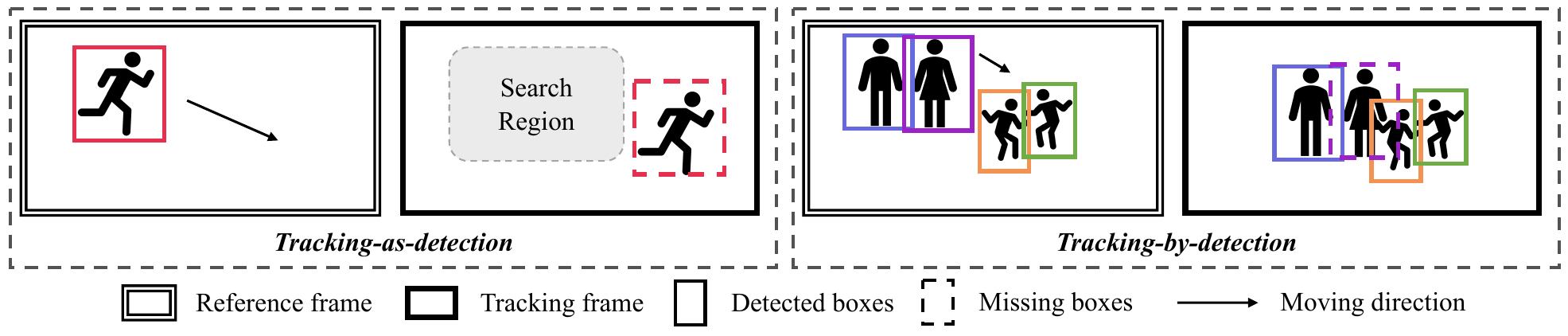}
   %\vspace{-0.25in}
   \caption{Reflection on previous tracking paradigms: the tracking-as-detection fails when the target object moves rapidly and the estimated search region is incorrect, while the tracking-by-detection fails when the target objects cannot be detected.}
   \label{fig:paradigm}
\end{figure*}

\subsection{Task-specific Tracking Models}
\label{subsec:tracking_models}
Over the years, a wide variety of task-specific tracking models~\cite{fan2019siamese,li2019evolution,oh2019video,cheng2021stcn,zeng2022motr,chu2023transmot,voigtlaender2019mots,porzi2020learning,wang2021end,yang2021crossover,wang2024omnivid} have been proposed to continuously improve the benchmark performance. Single object tracking (SOT)~\cite{smeulders2013visual}, also known as visual object tracking, requires trackers to estimate the location of an arbitrary object of interest in a video sequence. Siamese tracking methods~\cite{bertinetto2016fully,li2018high,li2019evolution,li2019siamrpn++,fan2019siamese,chen2020siamese,yu2020deformable,lin2021swintrack} have dominated this field for quite a long time, which train deep models to learn the matching between the target object and a search region cropped from the tracking frame. Video object segmentation (VOS) replaces the description of target objects in SOT with masks, which therefore relies on more fine-grained matching to obtain the per-pixel predictions. To this end, spatial-temporal memory networks~\cite{oh2019video,seong2021hierarchical,cheng2021mivos,cheng2021stcn,cheng2022xmem,wang2022look} have been widely explored to separate targeted objects
from the background through memory reading and updating in an online manner. 

Unlike the above-mentioned tracking of specified instances, another type of tracking task is dedicated to detecting all objects belonging to particular categories throughout a video and tracking their trajectories. Multiple object tracking (MOT)~\cite{berclaz2011multiple}, for example, aims to track all the pedestrians or vehicles on crowded streets. By far, tracking-by-detection~\cite{dicle2013way,bae2014robust,xiang2015learning,bergmann2019tracking,sun2020transtrack,zhang2021fairmot,liu2022online,zhang2022bytetrack} is the most popular solution for MOT, where all the frames are fed into a powerful detector~\cite{ge2021yolox} first and then the detected boxes are associated temporally according to their Intersection-of-Union (IoU) and appearance similarity. Multiple object tracking and segmentation (MOTS)~\cite{voigtlaender2019mots} additionally annotates instance masks on the MOT benchmark~\cite{milan2016mot16} to extend it to a more challenging problem. Due to the inherent correlation between both tasks, most MOTS models~\cite{voigtlaender2019mots,athar2020stem,wu2021track,meinhardt2022trackformer} are developed upon advanced MOT trackers by appending a mask head. Video instance segmentation (VIS)~\cite{yang2019video} shares the same task definition with MOTS but focuses on more open scenarios and more abundant categories. More precisely, it is an extension of instance segmentation~\cite{hafiz2020survey} in the video domain, which simultaneously performs detection, segmentation, and tracking of instances in videos collected from the YouTube~\cite{bertasius2020classifying,wang2021end,hwang2021video,yang2021crossover,wu2022defense}. 

\subsection{Unified Tracking Models}
\label{subsec:unified_tracking_models}
The customized tracking models developed in various tracking tasks lead to complicated training processes and poor generalization ability, which motivates a series of works~\cite{wang2019fast,wu2021track,wang2021different} exploring a unified architecture to solve multiple tracking tasks. UniTrack~\cite{wang2021different} argues that the core of object tracking lies in visual representation learning, based on which different tracking tasks could be resolved by propagation (\ie, SOT and VOS) or association (\ie, MOT, MOTS, and PoseTrack~\cite{andriluka2018posetrack}) between the learned features. To this end, they apply a shared appearance model and multiple parameter-free heads to address the above five tracking tasks within the same framework. Nevertheless, the remarkable discrepancies between different heads prevent UniTrack~\cite{wang2021different} from making use of large-scale tracking datasets for training and thus limiting its performance. UTT~\cite{ma2022unified} takes a further step by unifying SOT and MOT within a completely homogeneous architecture, which however does not support the per-pixel prediction tasks, \eg, VOS and MOTS. 

Comparatively, Unicorn~\cite{yan2022towards} achieves the grand unification of object tracking for the first time. It bridges the gap between different tracking tasks with a delicately designed target prior, which indicates the objects to be detected for the detection head. Despite the competitive performance on a wide variety of tracking benchmarks, there still exists some problems for Unicorn: 1) the inference pipelines for different tasks vary significantly, for example, they directly pick the box or mask with the highest confidence score as the final tracking result for SOT$\&$VOS, but implement the embedding association between detected boxes and existing trajectories for MOT$\&$MOTS; 2) they train the model on box tasks and mask tasks separately. To further promote the unification of training and inference processes of different tasks, the concurrent work UNINEXT~\cite{yan2023universal} proposes a novel prompt-based tracking framework, which perceives different types of tracking tasks through different prompts. In this paper, we reflect on the mainstream solutions developed for different tracking tasks and further present a tracking-with-detection paradigm. The proposed paradigm follows the spirit of tracking more closely and unifies various tracking tasks more simply. Through comprehensive experiments, we have validated the feasibility of unifying different tracking tasks and achieved consistent performance improvements. This confirms such a paradigm shift is of great value and potential to the research community.

\subsection{DETR-based Object Detectors}
\label{subsec:detr}
The development of object detection methods has seen a significant boost with the introduction of the DETR (DEtection TRansformers) architecture~\cite{carion2020end}. Since its inception, researchers have sought to enhance and refine the DETR model to achieve higher accuracy and versatility~\cite{zhu2021deformable,liu2022dab,li2022dn,zhang2022dino}. Deformable-DETR~\cite{zhu2021deformable} remarkably speeds up the convergence of DETR by attending to a small set of sampling points around a reference, instead of the whole image. DAB-DETR~\cite{liu2022dab}, on the other hand, utilizes the dynamic anchor boxes for the initialization of object queries. These modifications demonstrate the potential for refining the core components of DETR to address specific challenges in object detection. DN-DETR~\cite{li2022dn} takes a different approach by proposing a query-denoising technique. It contributes to the overall stability and reliability of the detection process, thereby augmenting the robustness of the DETR paradigm. In this paper, we build \system upon Deformable-DETR considering its simplicity and efficiency. We believe that the performance of our method will be further improved when equipped with more advanced detectors.

\begin{figure*}[t]
  \centering
   \includegraphics[width=\linewidth]{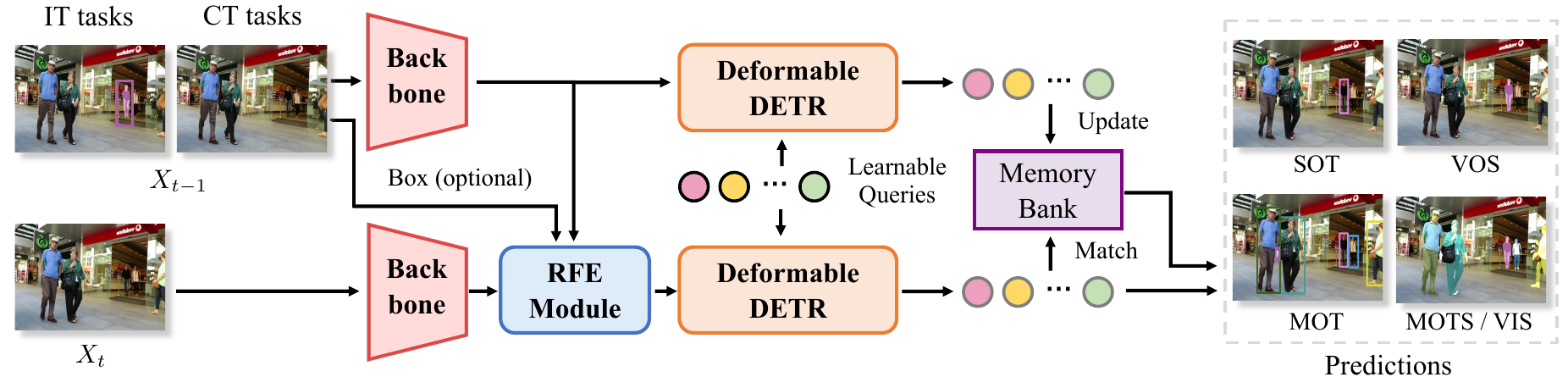}
   \caption{Overview of the proposed \system, which consists of a backbone network to extract the multi-scale frame features, a Reference-guided Feature Enhancement (REF) module to model the correlation between the target objects and the tracking frame, and a deformable DETR-based detector to predict the bounding boxes and instance masks. Note that we share the network architecture and inference pipeline for all the tracking tasks. IT: instance tracking, CT: category tracking.}
   \label{fig:network}
\end{figure*}

%-------------------------------------------------------------------------
\section{Method}
\label{sec:method}
In this paper, we propose a novel tracking-with-detection paradigm, where tracking and detection facilitate each other to enjoy the synergy. Based on this paradigm, we further introduce a unified tracking model, \system. In the following section, we start by reviewing the previous tracking paradigms that are developed by existing tracking models in Sec.~\ref{subsec:review}. Reflecting on their limitations, we then present our tracking-with-detection paradigm and illustrate its instantiation in Sec.~\ref{subsec:tracking-with-detection}. Finally, we introduce the loss functions for training and the unified online inference process in Sec.~\ref{subsec:loss} and Sec.~\ref{subsec:tracker}, respectively.

\subsection{A Brief Review on Previous Tracking Paradigms}
\label{subsec:review}
We categorize the existing tracking paradigms into two types: tracking-as-detection~\cite{bertinetto2016fully,li2018high,li2019siamrpn++,chen2020siamese,yu2020deformable,lin2021swintrack,oh2019video,cheng2021stcn,cheng2022xmem,wang2022look} and tracking-by-detection~\cite{bae2014robust,xiang2015learning,bergmann2019tracking,sun2020transtrack,zhang2021fairmot,liu2022online,zhang2022bytetrack,aharon2022bot,voigtlaender2019mots}. The former essentially treats tracking as a prior-guided detection problem, where the ``prior'' could be a template~\cite{fan2019siamese,cui2022mixformer,chen2023seqtrack} in the case of the SOT task or the memory bank~\cite{cheng2021stcn,cheng2022xmem} in the case of the VOS task. It typically involves a deep fusion between the template features (or memory) and the tracking frame, with the detection (or segmentation) head directly producing the bounding box (or mask) of the target object in the tracking frame. Notably, SOT techniques often reduce tracking computation by confining the detection to a smaller search region based on the historical object trajectory. In this paradigm, tracking unidirectionally provides reference information for detection, and thus the target object is difficult to track with the misestimated search region or false match with the memory (see Figure~\ref{fig:paradigm} left).

The latter, on the other hand, resolves the tracking task in a two-stage way, \ie,  first, predicting objects in various frames independently, and then establishing temporal associations. In particular, for the MOT methods~\cite{yu2022relationtrack,du2022strongsort,zhang2022bytetrack}, the classical practice is to fine-tune a detector on the tracking dataset to adapt it to a specific domain and assign the identity information to different boxes based on their IoU, as well as the similarity of corresponding feature embeddings. In this case, the detection completely ignores past tracking information.

In this paper, we argue that neither of the above approaches is fully compatible with the spirit of tracking. Unlike detecting isolated objects in single images, the information of the target object in the reference frame serves as a crucial guide for its identification in the current frame, which in turn could be leveraged for temporal correlation. In other words, tracking and detection should complement each other bidirectionally, thereby reinforcing the tracking process. With this in mind, we present a novel tracking paradigm, tracking-with-detection, for improved performance on both tracking tasks.

\subsection{Tracking-with-detection}
\label{subsec:tracking-with-detection}
The basic idea of our tracking-with-detection is to leverage tracking results to aid the detection of the whole tracking frame. To achieve this, we insert a Reference-guided feature Enhancement (RFE) module into a powerful detector~\cite{zhu2021deformable}, to supplement the detector with appearance priors obtained from previous tracking results. The overall framework of \system is illustrated in Figure~\ref{fig:network}.

Given a video sequence $\mathcal{V} = [\textit{X}_{1}, \textit{X}_{2}, ..., \textit{X}_{T}]$, tracking aims to estimate the location of moving objects over time. For the instance tracking tasks, the target objects are annotated in the first frame, in the form of bounding boxes $b_{0}$ = $
\{(x_{1}, y_{1}), (x_{2}, y_{2})\}$ for SOT, or instance masks $M_{0}$ for VOS. For category tracking tasks, \eg, MOT, MOTS, and VIS, all detected objects need to be tracked. Taking the tracking frame $\textit{X}_{t} \in \mathcal{R}^{3 \times H \times W}$ as inputs, we first adopt a backbone network to extract a pyramid of multi-scale features $F = \{f_{t}^{i}\}_{i=i}^{L}$, where $f_{t}^{i} \in \mathcal{R}^{C_{i} \times H/2^{i} \times W/2^{i}}$.

\subsubsection{Reference-guided Feature Enhancement} 
\label{subsubsec:rfe}
\begin{figure}[!ht]
\centering
\includegraphics[width=0.8\linewidth]{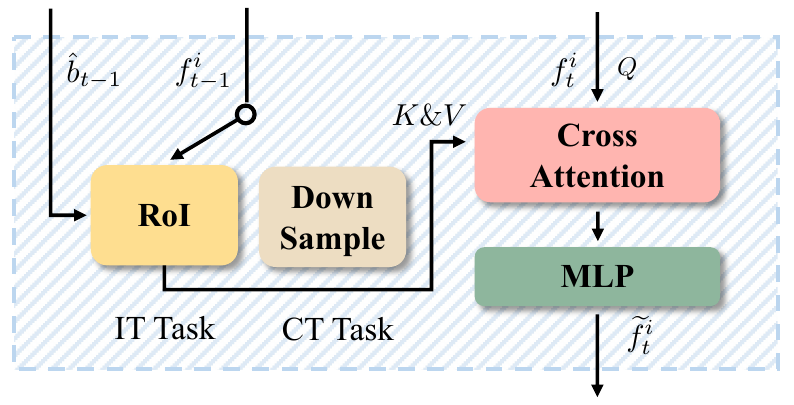}
\vspace{-0.15in}
\caption{Architecture of the proposed RFE module.}
\label{fig:rfe}
\end{figure}

Unlike object detection that localizes objects in a static image, tracking continuously updates the positions of target objects in a video sequence. This inspires us to supplement the backbone features of the tracking frame with the appearance priors from the previous tracking results, as illustrated in Figure~\ref{fig:rfe}. We discard the location priors here since the occlusions or movements of objects may result in a remarkable change of box coordinates between adjacent frames. 

More specifically, we enhance the feature pyramid of $\textit{X}_{t}$ with the RoIAlign~\cite{he2017mask} features of the target objects in $\textit{X}_{t-1}$ for the instance tracking tasks, while for the category tracking tasks, we instead downsample the feature map of $\textit{X}_{t-1}$ to provide the temporal contextual information:
\begin{equation}
h_{t-1}^{i}=\left\{
\begin{aligned}
\mathrm{RoIAlign}(f_{t-1}^{i}, \hat{b}_{t-1}) &, \text{instance tracking tasks} \\
\mathrm{DownSample}(f_{t-1}^{i}) &, \text{category tracking tasks}
\end{aligned}
\right.
\label{eq:roi_down}
\end{equation}

where $\hat{b}_{t-1}$ denote the tracked boxes in $\textit{X}_{t-1}$ and initialized with the ground-truth bounding boxes in the first frame\footnote{Note that for VOS, we calculate the bounding boxes by calling $\rm{torchvision.ops.masks\_to\_boxes}$. }. Taking $f_{t}^{i}$ as query, $h_{t-1}^{i}$ as key and value, we then model the correlation between them through cross-attention:
\begin{equation}
   g_{t}^{i} = \mathrm{CrossAttn}(f_{t}^{i}, h_{t-1}^{i}),
\end{equation}
where the normalization and flattening operations are omitted for simplicity. After that, $g_{t}^{i}$ is input to an MLP block to calculate the enhanced features $\widetilde{f}_{t}^{i}$. In this way, we obtain the refined feature pyramid $\widetilde{F} = \{\widetilde{f}_{t}^{i}\}_{i=i}^{L}$.

\subsubsection{Deformable DETR}
\label{subsubsec:ddetr}
We apply Deformable DETR~\cite{zhu2021deformable} as our object detector, which learns a set of object queries $q \in \mathcal{R}^{N \times C_{q}}$ to decode box coordinates and class labels. Specifically, taking $\widetilde{F}$ as input, a transformer encoder first exchanges the information between different scales through multi-scale deformable self-attention. After that, a transformer decoder is applied to interact the object queries with the image features through deformable cross-attention. Finally, we input the updated queries $\widetilde{q}_{t}$ to a 3-layer Feed-Forward Network (FFN) and a linear classification layer to produce the bounding boxes and category predictions, respectively. For the mask generation, we gather the multi-scale features from the transformer encoder through an FPN architecture~\cite{lin2017feature} to produce a high-resolution feature map $m_{t} \in \mathcal{R}^{C_{p} \times H/8 \times W/8}$ and employ $\widetilde{q}_{t}$ to generate instance-aware kernel weights $\omega$. With this, the instance masks $\hat{M}_{t}$ are predicted by a $1\times1$ convolution~\cite{tian2020conditional}:
\begin{equation}
    \hat{M}_{t} = \mathrm{CondConv_{1\times1}}(m_{t}, \omega),
\end{equation}

\subsection{Loss Functions}
\label{subsec:loss}
\subsubsection{Per-frame Detection Loss} 
We formulate object detection as a set prediction problem~\cite{carion2020end,zhu2021deformable} and resolve the label assignment from an optimal transport perspective~\cite{ge2021ota,wu2022defense}. We consider both classification accuracy and box GIoU~\cite{rezatofighi2019generalized} to compute the transport cost. In this way, each ground truth is assigned to $K$ predictions with the lowest costs, where $K$ is dynamically calculated for each sample. The overall per-frame detection loss is a weighted sum of multiple terms:
\begin{equation}
    \mathcal{L}_{det} = \mathcal{L}_{cls} + \lambda_{1} \mathcal{L}_{box} + \lambda_{2} \mathcal{L}_{mask},
\end{equation}
where a binary cross-entropy loss is adopted as $\mathcal{L}_{cls}$,  $\mathcal{L}_{box}$ is a combination of L1 loss and the GIoU loss, and $\mathcal{L}_{mask}$ is a combination of the dice loss~\cite{milletari2016v} and the focal loss~\cite{lin2017focal}. We set $\lambda_{1}$ and $\lambda_{2}$ to 2.0 by default.

\subsubsection{Contrastive ReID Loss} 
Several works~\cite{yao2021efficient,wang2022anchor} reveal that the interaction with local image features empowers the learnable queries in DETR-like object detectors~\cite{carion2020end,zhu2021deformable} with both appearance and positional information. To this end, we combine these queries with the corresponding RoIAlign~\cite{he2017mask} features for the instance-level association between different frames. For the $k_{th}$ query $\widetilde{q}_{t}^{k}$, we first extract its RoIAlign features $r_{t}^{k}$ from $f_{t}^{L}$, the backbone feature with a size of $H/8 \times W/8$, and then send them to an MLP block to generate an identity embedding $e_{t}^{k}$:
\begin{equation}
    e_{t}^{k} = \mathrm{MLP}(\mathrm{concat}(\widetilde{q}_{t}^{k}, r_{t}^{k})),
\end{equation}
During training, we randomly sample a reference frame from the same video, and learn the discriminative identity embedding in a contrastive manner:
\begin{equation}
    \mathcal{L}_{reid} = \mathbf{log}[1 + \sum_{e_{ref}^{+}}\sum_{e_{ref}^{-}}\mathrm{exp}(e_{t} \cdot e_{ref}^{-} - e_{t} \cdot e_{ref}^{+})]. 
\end{equation}
where $e_{ref}$ denotes the identity embeddings in the reference frame. We select $p$ predictions with the least cost as positives and $n$ predictions with the highest costs as negatives. Finally, our model is supervised with $\mathcal{L}_{det}$ + $\beta$ $\mathcal{L}_{reid}$, where $\beta$ is a trade-off parameter and set to 10 empirically.

\subsection{Unified Online Tracking}
\label{subsec:tracker}
In contrast to Unicorn~\cite{yan2022towards}, we adopt the same tracking pipeline for different tasks during inference, where the localization and appearance information are combined for the association between the detected boxes and existing trajectories. Specifically, we maintain a memory bank for each trajectory, which stores the historical identity embeddings to utilize the temporal information for more robust matching. Specifically, assume there are $N$ detected objects in the $t_{th}$ frame and $M$ trajectories with the corresponding memory banks $\{e_{k}^{m}\}_{k=1}^{t-1}$, for $m=1..M$\footnote{For SOT, $M$ is always 1, and for VOS, $M$ is equal to the number of instances annotated in the first frame.}, we first calculate the temporally weighted sum of the identity embeddings stored in each memory bank: $\widetilde{e}^{m} = \sum_{k=1}^{t-1} \frac{k}{t-1} \times e_{k}^{m}$, and then compute the bi-directional similarity between $\{\widetilde{e}^{m}\}_{m=1}^{M}$ and the identity embeddings of the detected instances:
\begin{equation}
    s_{n, m} = [\frac{\mathrm{exp}(\widetilde{e}^{m} \cdot e_{t}^{n})}{\sum_{h=1}^{N}\mathrm{exp}(\widetilde{e}^{m} \cdot e_{t}^{h})} + \frac{\mathrm{exp}(\widetilde{e}^{m} \cdot e_{t}^{n})}{\sum_{h=1}^{M}\mathrm{exp}(\widetilde{e}^{h} \cdot e_{t}^{n})} ] / 2, 
\end{equation}
where $e_{t}^{n}$ is the identity embedding of the $n_{th}$ instance in the $t_{th}$ frame. 

We also adopt Kalman filter~\cite{brown1997introduction} as the motion model to keep track of the detected objects and predict their locations in the tracking frame~\cite{zhang2021fairmot,zhang2022bytetrack,aharon2022bot}. With this, we could calculate the IoU between the $N$ detected bounding boxes and the estimated locations of $M$ trajectories by the Kalman filter. Then the boxes with low IoU are filtered out: 
\begin{equation}
\begin{aligned}
    M_{iou} &= \rm{IoU}  <  \tau, \\
    S &= S[M_{iou}].
\end{aligned}
\end{equation}
where $\tau$ is set to 0.25 empirically. Finally, taking $1 - S$ as the cost matrix, we resolve the box assignment problem with the Hungarian algorithm~\cite{kuhn1955hungarian}. The unmatched boxes whose detection scores are larger than $\tau_{new}$ will be initialized as a new track, and we set $\tau_{new} = 1.0$ for the instance tracking tasks.

%-------------------------------------------------------------------------
\section{Experiments}
\label{sec:exp}

\subsection{Implementation Details}
\noindent \textbf{Training.} The complete training process consists of three stages: during the first stage, we pre-train the model on COCO~\cite{lin2014microsoft} for object detection and instance segmentation following Unicorn~\cite{yan2022towards}. Then we follow the common practice in VIS methods~\cite{hwang2021video,wu2022seqformer,wu2022defense} to randomly crop the same image from COCO twice to form a pseudo key-reference frame pair. Finally, the proposed \system is fine-tuned on the training splits of various tracking datasets, during which two images are randomly sampled as key and reference frames respectively. Note that we treat COCO~\cite{lin2014microsoft} as an additional downstream dataset and perform joint training on it to make the training process more stable. 

\begin{figure}[!ht]
  \centering
   \includegraphics[width=\linewidth]{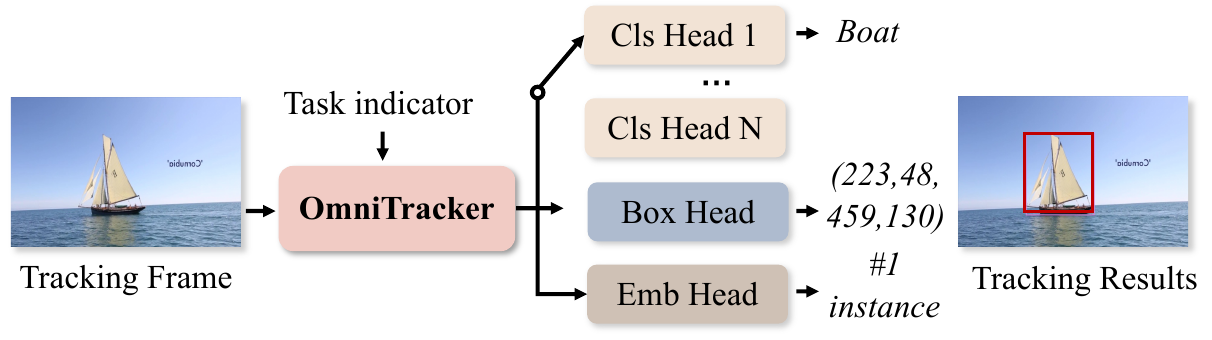}
   \caption{Illustration of different heads in \system. All components marked with solid lines are shared for different tasks, while components marked with dotted lines are task-specific. We input a task indicator to the model to indicate which classifier to use during both training and inference.}
   \label{fig:head}
\end{figure}

We implement 6 data loaders for COCO, SOT (TrackingNet~\cite{muller2018trackingnet}, LaSOT~\cite{fan2019lasot}, GOT10K~\cite{huang2019got}), VOS (DAVIS17~\cite{pont20172017}), MOT (MOT17~\cite{milan2016mot16}), MOTS (MOTS20~\cite{voigtlaender2019mots}), and VIS (YTVIS19~\cite{yang2019video}), and iteratively feed one of them to the model for joint training. We use different classifiers for different tasks and set the category annotation of SOT and VOS to 0. As mentioned in Sec.~\ref{subsubsec:ddetr}, we employ a classification head to predict the categories of objects, a box head to perform bounding box regression, and an embedding head for the cross-frame association. We share all the heads other than the classification head between different tasks to promote parameter efficiency and facilitate knowledge transfer. Multi-scale training is applied and the shortest side ranges from 736 to 864, following~\cite{yan2022towards}. The batch sizes for the three stages are 32, 32, and 16, and the training iterations are 185k, 148k, and 144k. The parameters are optimized with AdamW~\cite{loshchilov2018decoupled}. The initial learning rates are all set to $1e^{-4}$ for three stages, and decayed by 0.1 after 148k, 59.2k, and 8k iterations. The RoI size and downsampled size in Equation~\ref{eq:roi_down} are both set to 7. We adopt Swin Transformer~\cite{liu2021swin}-Tiny and Swin Transformer-Large as backbones in the main experiments, and Swin Transformer-Tiny as the backbone in the ablation. 

UNINEXT~\cite{yan2023universal} adopts a more advanced detector DINO~\cite{zhang2022dino} than \system, and first pre-train the model on a large-scale object detection dataset Objects365~\cite{shao2019objects365} for learning universal knowledge about objects. For a fair comparison, we skip the Objects365 pretraining stage and abandon the Box De-noising strategy, the results are denoted as UNINEXT-L-noObjPre. In addition, since Unicorn~\cite{yan2022towards} and UNINEXT~\cite{yan2023universal} only report the results with ConvNeXt~\cite{liu2022convnet}-Large as the backbone, we also reproduce their methods with Swin Transformer-Tiny~\cite{liu2021swin} as the backbone for a fair comparison, which is denoted as Unicorn-T and UNINEXT-T-noObjPre. Note that we conduct ablation studies on the validation split of MOT17 for MOT unless otherwise mentioned.

\begin{table}[t]
\centering
\caption{Comparisons with the SOT-specific trackers and unified trackers on LaSOT~\cite{fan2019lasot} and TrackingNet~\cite{muller2018trackingnet}. }
\label{tab:sot}
\renewcommand\arraystretch{1.1}
\setlength{\tabcolsep}{0pt} % let TeX compute the intercolumn space
\begin{tabular*}{\linewidth}{@{\extracolsep{\fill}}lc | cccc | cccc @{}}
\toprule
\small
\multirow{2}*{\textbf{Method}} && \multicolumn{3}{c}{\textbf{LaSOT}} && \multicolumn{3}{c}{\textbf{TrackingNet}}  \\
~ && Suc & P$_{norm}$ & P && Suc & P$_{norm}$ & P \\
\midrule
SiamFC~\cite{bertinetto2016fully} && 33.6 & 42.0 & 33.9 && 57.1 & 66.3 & 53.3 \\
D3S~\cite{lukezic2020d3s} && - & - & - && 72.8 & 76.8 & 66.4 \\
ATOM~\cite{danelljan2019atom} && 51.5 & 57.6 & 50.5 && 70.3 & 77.1 & 64.8 \\
SiamPRN++~\cite{li2019siamrpn++} && 49.6 & 56.9 & 49.1 && 73.3 & 80.0 & 69.4 \\
DiMP~\cite{bhat2019learning} && 56.9 & 65.0 & 56.7 && 74.0 & 80.1 & 68.7 \\
KYS~\cite{bhat2020know} && 55.4 & 63.3 & - && 74.0 & 80.0 & 68.8 \\
Ocean~\cite{zhang2020ocean} && 56.0 & 65.1 & 56.6 && - & - & - \\
AutoMatch~\cite{zhang2021learn} && 58.2 & - & 59.9 && 76.0 & - & 72.6 \\
PrDiMP~\cite{danelljan2020probabilistic} && 59.8 & 68.8 & 60.8 && 75.8 & 81.6 & 70.4 \\
TrDiMP~\cite{wang2021transformer} && 63.9 & - & 61.4 && 78.4 & 83.3 & 73.1 \\
Siam R-CNN~\cite{voigtlaender2020siam} && 64.8 & 72.2 & - && 81.2 & 85.4 & 80.0 \\
TransT~\cite{chen2021transformer} && 64.9 & 73.8 & 69.0 && 81.4 & 86.7 & 80.3 \\
KeepTrack~\cite{mayer2021learning} && 67.1 & 77.2 & 70.2 && - & - & -\\
STARK~\cite{yan2021learning} && 67.1 & 77.0 & - && 82.0 & 86.9 & - \\
SwinTrack~\cite{lin2021swintrack} &&  67.2 & - & 70.8 && 81.1 & - & 78.4 \\
AiATrack~\cite{gao2022aiatrack} && - & 79.4 & 73.8 && - & 87.8 & 80.4 \\
OSTrack~\cite{ye2022joint} && - & 78.7 & 75.2 && - & 87.8 & 82.0 \\
SimTrack-L~\cite{chen2022backbone} && - & 79.7 & - && - & 87.4 & - \\
MixFormer-L~\cite{cui2022mixformer} && - & 79.9 & 76.3 && - & 88.9 & 83.1 \\
OmniVid~\cite{wang2024omnivid} && 70.8 & 79.6 & 76.9 && 83.8 & 88.9 & 83.2 \\
DropTrack~\cite{wu2023dropmae} && 71.8 & 81.8 & 78.1 && 84.1 & 88.9 & - \\
GRM-L~\cite{gao2023generalized} && 71.4 & 81.2 & 77.9 && 84.4 & 88.9 & 84.0 \\
TATrack-L~\cite{he2023target} && 71.1 & 79.1 & 76.1 && 85.0 & 89.3 & 84.5 \\
SeqTrack-L~\cite{chen2023seqtrack} && 72.1 & 81.7 & 79.0 && 85.0 & 89.5 & 84.9  \\
HIPTrack-L~\cite{cai2024hiptrack} && 72.7 & 82.9 & 79.5 && 77.4 & 88.0 & 74.5 \\
ARTrack-L~\cite{wei2023autoregressive} && 73.1 & 82.2 & 80.3 && 85.6 & 89.6 & 86.0 \\
ARTrackV2-L~\cite{bai2024artrackv2} && 73.6 & 82.8 & 81.1 && 86.1 & 90.4 & 86.2 \\
ODTrack-L~\cite{zheng2024odtrack} && 74.0 & 84.2 & 82.3 && 86.1 & 91.0 & 86.7 \\
SAMURAI-L~\cite{yang2024samurai} && 74.2 & 82.7 & 80.2 && - & - & - \\
LoRAT-L~\cite{lin2025tracking} && 74.2 & 83.6 & 80.9 && 85.0 & 89.5 & 84.4 \\
\midrule
UniTrack~\cite{wang2021different} && 35.1 & - & 32.6 && - & - & - \\
UTT~\cite{ma2022unified} && 64.6 & - & 67.2 && 79.7 & - & 77.0 \\
 Unicorn-T~\cite{yan2022towards} && 65.3 & 73.1 & 68.7 && 79.0 & 82.0 & 77.4 \\
 UNINEXT-T-noObjPre~\cite{yan2023universal} && 66.2 & 73.7 & 69.5 && 79.7 & 82.1 & 77.5 \\
\system-T && 65.9 & 73.5 & 69.3 && 80.2 & 82.5 & 77.7 \\
Unicorn-L~\cite{yan2022towards} && 68.5 & 76.6 & 74.1 && 83.0 & 86.4 &  82.2 \\
UNINEXT-L~\cite{yan2023universal} && 72.4 & 80.7 & 78.9 && 85.1 & 88.2 & 84.7 \\
UNINEXT-L-noObjPre && 69.5 & 78.1 & 75.8 && 83.2 & 86.6 & 82.3  \\
\system-L && 69.1 & 77.3 & 75.4 && 83.4 & 86.7 & 82.3 \\
\bottomrule
\end{tabular*}
\end{table}

\noindent \textbf{Inference.} During inference, we implement the memory bank mentioned in Sec.~\ref{subsec:tracker} as a First-In-First-Out (FIFO) queue. Note that for the instance tracking tasks, we store the identity embedding of the box whose IoU with the ground truth box is highest in the memory for the first frame, and always reserve it during memory update. The memory size is set to 64, 64, 32, and 3 for SOT, VOS, MOT(S), and VIS. 

Next, we report the performance of \system on various tracking tasks.

\subsection{Evaluation on the Instance Tracking Tasks}

\subsubsection{Single Object Tracking}
In Table~\ref{tab:sot}, a comprehensive evaluation of \system is presented, comparing its performance against both SOT-specific and unified tracking models on two prominent and extensive SOT benchmarks: LaSOT~\cite{fan2019lasot} and TrackingNet~\cite{muller2018trackingnet}. The LaSOT dataset comprises 1,120 training sequences and 280 testing sequences, while the TrackingNet dataset includes a training set of 30,000 sequences and a testing set containing 511 sequences. This evaluation is based on key performance metrics including Success (Suc), precision (P), and normalized precision (P$_{norm}$).

It can be seen that although UNINEXT~\cite{yan2023universal} adopts a stronger object detector than \system, \ie, DINO~\cite{zhang2022dino} $v.s.$ Deformable-DETR~\cite{zhu2021deformable}, \system achieves more superior performance than UNINEXT-T-noObjPre~\cite{yan2023universal} on TrackingNet when utilizing Swin-Tiny as the backbone network. We believe that the slight performance gap on LaSOT can be bridged if we utilize the same detector as UNINEXT. While compared to Unicorn, our approach demonstrates an obvious improvement on both LaSOT and TrackingNet datasets, \ie, 0.4\% and 0.5\% in terms of P$_{norm}$, respectively. Importantly, even when transitioning to a larger backbone architecture, we maintain a competitive advantage over the majority of unified models, while also achieving commendable results in comparison to the SOT-specific models. This performance comparison underscores the efficacy of our method in achieving state-of-the-art results across diverse benchmark datasets. 

\begin{table}[t]
\centering
\caption{Comparisons with the VOS-specific trackers and unified trackers on DAVIS 2016 and 2017 val~\cite{pont20172017}. \textbf{Mem} indicates whether a memory bank is maintained. }
\label{tab:vos}
  \renewcommand\arraystretch{1.1}
  \setlength{\tabcolsep}{0.7pt} % let TeX compute the intercolumn space
  \begin{tabular*}{\linewidth}{@{\extracolsep{\fill}}lcc | cccc | cccc @{}}
\toprule
\small
\multirow{2}*{\textbf{Method}} & \multirow{2}*{\textbf{Mem}} && \multicolumn{3}{c}{\textbf{DAVIS16}} && \multicolumn{3}{c}{\textbf{DAVIS17}}  \\
~ & ~ && $\mathcal{J\&F}$ & $\mathcal{J}$ & $\mathcal{F}$ && $\mathcal{J\&F}$ & $\mathcal{J}$ & $\mathcal{F}$ \\
\midrule
FAVOS~\cite{cheng2018fast} & \XSolidBrush && 81.0 & 82.4 & 79.5 && 58.2 & 54.6 & 61.8 \\
OSMN~\cite{maninis2018video} & \XSolidBrush && 73.5 & 74.0 & 72.9 && 54.8 & 52.5 & 57.1 \\
VideoMatch~\cite{hu2018videomatch} & \XSolidBrush && - & 81.0 & - && 56.5 & - & - \\
RANet~\cite{wang2019ranet} & \XSolidBrush && 85.5 & 85.5 & 85.4 && 65.7 & 63.2 & 68.2 \\
FRTM~\cite{robinson2020learning} & \Checkmark && 83.5 & 83.6 & 83.4 && 76.7 & 73.9 & 79.6 \\
LWL~\cite{bhat2020learning} & \Checkmark && - &  - & - && 81.6 & 79.1 & 84.1 \\
STM~\cite{oh2019video} & \Checkmark && 89.3 & 88.7 & 89.9 && 81.8 & 79.2 & 84.3 \\
CFBI~\cite{yang2020collaborative} & \Checkmark && 89.4 & 88.3 & 90.5 && 81.9 & 79.1 & 84.6 \\
HMMN~\cite{seong2021hierarchical} & \Checkmark && 90.8 & 89.6 & 92.0 && 84.7 & 81.9 & 87.5 \\
STCN~\cite{cheng2021stcn} & \Checkmark && 91.6 & 90.8 & 92.5 &&  85.4 & 82.2 & 88.6 \\
XMem~\cite{cheng2022xmem} & \Checkmark && 92.0 & 90.7 & 93.2 && 87.7 & 84.0 & 91.4 \\
ISVOS~\cite{wang2022look} & \Checkmark && 92.8 & 91.8 & 93.8 && 88.2 & 84.5 & 91.9 \\
DEVA~\cite{cheng2023tracking} & \Checkmark && - & - & - && 86.8 & 83.6 & 90.0 \\
Cutie~\cite{cheng2024putting} & \Checkmark && - & - & - && 88.8 & 85.4 & 92.3 \\
\midrule
SiamMask~\cite{wang2019fast} & \XSolidBrush && 69.8 & 71.7 & 67.8 && 56.4 & 54.3 & 58.5 \\
D3S~\cite{lukezic2020d3s} & \XSolidBrush && 74.0 & 75.4 & 72.6 && 60.8 & 57.8 & 63.8 \\
Siam R-CNN~\cite{voigtlaender2020siam} & \XSolidBrush && - & - & - && 70.6 & 66.1 & 75.0 \\
\midrule
UniTrack~\cite{wang2021different} & \Checkmark && - & - & - && - & 58.4 & - \\
Unicorn-T~\cite{yan2022towards} & \XSolidBrush && 83.2 & 83.0 & 83.4 && 64.5 & 62.7 & 66.3 \\
UNINEXT-T-noObjPre~\cite{yan2023universal} & \Checkmark && - & - & - && 65.5 & 63.8 & 67.2 \\
\system-T & \Checkmark && 84.7 & 84.1 & 85.3 && 66.2 & 64.9 & 67.5 \\
Unicorn-L~\cite{yan2022towards} & \XSolidBrush && 87.4 & 86.5 & 88.2 && 69.2 & 65.2 & 73.2 \\
UNINEXT-L~\cite{yan2023universal} & \Checkmark && - & - & - && 77.2 & 73.2 & 81.2 \\
UNINEXT-L-noObjPre & \Checkmark && - & - & - && 73.2 & 69.7 & 76.7 \\
\system-L & \Checkmark && 88.5 & 87.3 & 89.7 && 71.0 & 66.8 & 75.2 \\
\bottomrule
\end{tabular*}
\end{table}

\subsubsection{Video Object Segmentation}
When transitioning from single-object tracking (SOT) to video object segmentation (VOS), the challenges become even more intricate, as VOS involves a per-pixel classification task that demands capturing fine-grained information. In this context, we conduct an extensive evaluation of \system on both the single-object VOS benchmark, DAVIS 2016~\cite{pont20172017}, and its multi-object extension, DAVIS 2017. Our evaluation employs the following metrics: the mean Jaccard index ($\mathcal{J}$), the mean boundary $\mathcal{F}$ score, and their average, denoted as $\mathcal{J\&F}$.

To provide a comprehensive benchmark, we compare our results with those of several multi-task instance trackers, such as SiamMask~\cite{wang2019fast}, D3S~\cite{lukezic2020d3s}, and Siam R-CNN~\cite{voigtlaender2020siam}, which serve as relevant points of reference. Compared to them, our approach exhibits a substantial margin of improvement across both DAVIS 2016 and 2017 datasets. The results indicate that a simple multi-task paradigm will not lead to satisfactory performance on each task. We also present the performance comparison with other unified tracking models, including UniTrack~\cite{wang2021different}, Unicorn~\cite{yan2022towards}, and UNINEXT~\cite{yan2023universal}. The results demonstrate that our method outperforms UniTrack and Unicorn by a clear margin using models of different sizes. In terms of the $\mathcal{J\&F}$ metric, \system-L outperforms Unicorn-L by 1.1\% on DAVIS 2016 and 1.8\% on DAVIS 2017, thereby underscoring the superior segmentation accuracy of the OmniTracker. As for UNINEXT, with smaller models, our method performs better than UNINEXT since the performance gap of the detector itself is not significant. When larger models are adopted, the performance of the detector dominates the performance on the VOS task, so our approach falls slightly short compared to UNINEXT.

While there may still be a discernible gap between our approach (as well as other multi-task and unified tracking models) and task-specific VOS methods, it is important to emphasize that \system excels in simultaneously addressing multiple tracking tasks. This inherent versatility positions our approach as a powerful solution, offering a harmonious blend of flexibility and efficiency.

\begin{table}[t]
\centering
\caption{Comparisons with the MOT-specific trackers and unified trackers on  MOT17~\cite{milan2016mot16} test set. }
\label{tab:mot}
\renewcommand\arraystretch{1.1}
\footnotesize
\setlength{\tabcolsep}{0.7pt} % let TeX compute the intercolumn space
\begin{tabular*}{\linewidth}{@{\extracolsep{\fill}}lc | ccccccc @{}}
\toprule
\small
\textbf{Method} && MOTA$\uparrow$ & IDF1$\uparrow$ & HOTA$\uparrow$ & FP$\downarrow$ & FN$\downarrow$ & IDs$\downarrow$ \\
\midrule
TransCenter~\cite{xu2021transcenter} && 73.2 & 62.2 & 54.5 & 23112 & 123738 & 4614 \\
FairMOT~\cite{zhang2021fairmot} && 73.7 & 72.3 & 59.3 & 27507 & 117477 & 3303 \\
RelationTrack~\cite{yu2022relationtrack} && 73.8 & 74.7 & 61.0 & 27999 & 118623 & 1374 \\
CSTrack~\cite{liang2022rethinking} && 74.9 & 72.6 & 59.3 & 23847 & 114303 & 3567 \\
TransTrack~\cite{sun2020transtrack} && 75.2 & 63.5 & 54.1 & 50157 & 86442 & 3603 \\
SiamMOT~\cite{liang2022one} && 76.3 & 72.3 & - & - & - & - \\
CorrTracker~\cite{wang2021multiple} && 76.5 & 73.6 & 60.7 & 29808 & 99510 & 3369 \\
TransMOT~\cite{chu2023transmot} &&  76.7 & 75.1 & 61.7 & 36231 & 93150 & 2346 \\
MAATrack~\cite{stadler2022modelling} && 79.4 & 75.9 & 62.0 & 37320 & 77661 & 1452 \\
OCSORT~\cite{cao2022observation} && 78.0 & 77.5 & 63.2 & 15129 & 107055 & 1950 \\
MOTRV2~\cite{zhang2023motrv2} && 78.6 & 75.0 & 62.0 & 23409 & 94797 & 2619 \\
StrongSORT++~\cite{du2022strongsort} && 79.6 & 79.5 & 64.4 & 27876 & 86205 & 1194 \\
ByteTrack~\cite{zhang2022bytetrack} && 80.3 & 77.3 & 63.1 & 25491 & 83721 & 2196 \\
Bot-SORT~\cite{aharon2022bot} && 80.5 & 80.2 & 65.0 & 22521 & 86037 & 1212 \\
UCMCTrack~\cite{yi2024ucmctrack} && 79.0 & 79.0 & 64.3 & - & - & - \\
Deep OC-SORT~\cite{maggiolino2023deep} && 79.4 & 80.6 & 64.9 & 16600 & 98800 & 1023 \\
FeatureSORT~\cite{hashempoor2024featuresort} && 80.6 & 76.7 & 64.2 & 27581 & 84362 & 2637 \\
BoostTrack++~\cite{stanojevic2024boosttrack++} && 80.7 & 82.2 & 66.6 & - & - & - \\
SparseTrack~\cite{liu2023sparsetrack} && \textcolor{red}
{81.0} & 80.1 & 65.1 & 23904 & 81927 & 1170 \\
C-BIoU~\cite{yang2023hard} && 81.1 & 79.7 & 64.1 & - & - & -\\
MotionTrack~\cite{qin2023motiontrack} && 81.1 & 80.1 & 65.1 & 23802 & 81660 & 1140 \\
SMILEtrack~\cite{wang2024smiletrack} && 81.1 & 80.5 & 65.3 & 22963 & 79428 & 1246 \\
\midrule
Unicorn-T~\cite{yan2022towards} && 73.1 & 73.3 & 60.1 & 28434 & 121245 & 2094 \\
\system-T && 73.2 & 75.4 & 61.0 & 30216 & 97914 & 2367 \\
Unicorn-L~\cite{yan2022towards} && 77.2 & 75.5 & 61.7 & 50087 & 73349 & 5379 \\
\system-L && 79.1 & 75.6 & 62.3 & 29247 & 87192 & 1968 \\
\bottomrule
\end{tabular*}
\end{table}

\subsection{Evaluation on Category Tracking Tasks}

\subsubsection{Multiple Object Tracking}
We perform multiple object tracking on the most popular dataset, MOT17~\cite{milan2016mot16}, which focuses on pedestrian tracking in crowded scenes. Comprising 7 sequences each in both the training and test sets, MOT17 provides a comprehensive benchmark for assessing the capabilities of tracking models in complex scenarios. Six representative metrics are reported for quantitative comparison, including Multiple-Object Tracking Accuracy (MOTA), Identity F1 Score (IDF1), False Positives (FP), False Negatives (FN), and Identity Switches (IDS). Note that UNINEXT~\cite{yan2023universal} does not report the results on MOT17 and MOTS. 

The results in Table~\ref{tab:mot} show that \system achieves 79.1\% and 75.6\% in terms of MOTA and IDF1, surpassing Unicorn~\cite{yan2022towards} by 1.9\% and 0.1\%, respectively. This performance underscores the robustness and efficacy of our approach in addressing complex, real-world tracking scenarios with a multitude of pedestrians. Moreover, it is important to acknowledge that several state-of-the-art MOT-specific tracking models, such as ByteTrack~\cite{zhang2022bytetrack}, bolster their training data with Cityperson~\cite{zhang2017citypersons} and ETHZ~\cite{ess2008mobile} datasets. By incorporating such additional training data, we are confident that our model's performance will experience further augmentation.

\subsubsection{Multiple Object Tracking and Segmentation}
To elevate the complexity of multiple object tracking, the MOTS20 dataset~\cite{voigtlaender2019mots} offers a more formidable challenge by augmenting the MOT17 dataset~\cite{milan2016mot16} with pixel-wise annotations. Within this context, we present a comprehensive evaluation of \system on the MOTS20 dataset, utilizing a set of pertinent evaluation metrics, including the segment-based Multiple Object Tracking and Segmentation Accuracy (sMOTSA), Identity F1 Score (IDF1), False Positives (FP), False Negatives (FN), and Identity Switches (IDS). Note that sMOTSA differs from MOTA in that it is based on the mask overlap. 

Notably, our approach outperforms PointTrackV2~\cite{xu2021segment} and Unicorn~\cite{yan2022towards} by noteworthy margins of 5.2\% and 2.2\% in terms of sMOTSA, respectively. The results underscore the effectiveness of our method in addressing the complexities of multi-object tracking with segmentation considerations. 

\begin{table}[t]
\centering
\caption{Comparisons with the MOTS-specific approaches and unified trackers on  MOTS20~\cite{milan2016mot16}. }
\label{tab:mots}
\renewcommand\arraystretch{1.1}
\small
\setlength{\tabcolsep}{0.pt} % let TeX compute the intercolumn space
\begin{tabular*}{\linewidth}{@{\extracolsep{\fill}}lc | ccccccc @{}}
\toprule
\textbf{Method} && sMOTSA$\uparrow$ & IDF1$\uparrow$ & FP$\downarrow$ & FN$\downarrow$ & IDs$\downarrow$ \\
\midrule
Track R-CNN~\cite{voigtlaender2019mots} &&  40.6 & 42.4 & 1261 & 12641 & 567 \\
TraDeS~\cite{wu2021track} &&  50.8 & 58.7 & 1474 & 9169 & 492 \\
TrackFormer~\cite{meinhardt2022trackformer} && 54.9 & 63.6 & 2233 & 7195 & 278\\
PointTrackV2~\cite{xu2021segment} && 62.3 & 42.9 & 963 & 5993 & 541 \\
\midrule
Unicorn-T~\cite{yan2022towards} && 61.5 & 60.8 & 2675 & 6509 & 466 \\
\system-T && 62.2 & 62.5 & 1394 & 5805 & 411 \\
Unicorn-L~\cite{yan2022towards} && 65.3 & 65.9 & 1364 & 4452 & 398 \\
\system-L && 67.5 & 69.2 & 780 & 5204 & 215 \\
\bottomrule
\end{tabular*}
\end{table}

\begin{table}[t]
\centering
\caption{Comparisons with the VIS-specific trackers on YouTube-VIS~\cite{yang2019video} 2019 val set. Currently, there are no unified tracking models that support the VIS task. }
\label{tab:vis}
\renewcommand\arraystretch{1.1}
\setlength{\tabcolsep}{0.pt} % let TeX compute the intercolumn space
\begin{tabular*}{\linewidth}{@{\extracolsep{\fill}}lc | cccccc @{}}
\toprule
\small
\textbf{Method} && mAP & AP50 & AP75 & AR1 & AR10 \\
\midrule
MaskTrack R-CNN~\cite{yang2019video} && 31.8 & 53.0 & 33.6 & 33.2 & 37.6 \\
CrossVIS~\cite{yang2021crossover} && 36.6 & 57.3 & 39.7 & 36.0 & 42.0 \\
PCAN~\cite{ke2021prototypical} &&  37.6 & 57.2 & 41.3 & 37.2 & 43.9 \\
STEm-Seg~\cite{athar2020stem} && 34.6 & 55.8 & 37.9 & 34.4 & 41.6 \\
VisTR~\cite{wang2021end} && 40.1 & 64.0 & 45.0 & 38.3 & 44.9 \\
MaskProp~\cite{bertasius2020classifying} && 42.5 & - & 45.6 & - & - \\
Propose-Reduce~\cite{lin2021video} && 43.8 & 65.5 & 47.4 & 43.0 & 53.2 \\
IFC~\cite{hwang2021video} && 44.6 & 69.2 & 49.5 & 44.0 & 52.1 \\
SeqFormer-Res101~\cite{wu2022seqformer} && 49.0 & 71.1 & 55.7 & 46.8 & 56.9 \\
IDOL-Res101~\cite{wu2022defense} && 50.1 & 73.1 & 56.1 & 47.0 & 57.9 \\ 
SeqFormer-L~\cite{wu2022seqformer} && 59.3 & 82.1 & 66.4 & 51.7 & 64.4 \\
MinVIS-L~\cite{huang2022minvis} && 61.6 & 83.3 & 68.6 & 54.8 & 66.6 \\
VITA-L~\cite{heo2022vita} && 63.0 & 86.9 & 67.9 & 56.3 & 68.1 \\
IDOL-L$^{*}$~\cite{wu2022defense} && 63.0 & 87.0 & 69.7 & 55.0 & 67.9 \\
\textcolor{gray}{IDOL-L}~\cite{wu2022defense} && \textcolor{gray}{64.3} & \textcolor{gray}{87.5} & \textcolor{gray}{71.0} & \textcolor{gray}{55.6} & \textcolor{gray}{69.1} \\
GenVIS-L~\cite{heo2023generalized} && 64.0 & 84.9 & 68.3 & 56.1 & 69.4 \\
DVIS-L~\cite{zhang2023dvis} && 64.9 & 88.0 & 72.7 & 56.5 & 70.3 \\
DVIS-DAQ-L~\cite{zhou2024improving} && 65.7 & - & 73.6 & - & 70.7 \\
NOVIS-L~\cite{meinhardt2023novis} && 65.7 & 87.8 & 72.2 & 56.3 & 70.3 \\
\midrule
UNINEXT-T-noObjPre~\cite{yan2023universal} && 58.7 & 84.0 & 63.9 & 52.8 & 65.0 \\
\system-T && 58.8 & 82.4 & 64.9 & 52.4 & 65.5 \\
GLEE-L~\cite{wu2024general} && 63.6 & 85.2 & 70.5 & - & - \\
UNINEXT-L~\cite{yan2023universal} && 64.3 & 87.2 & 71.7 & - & - \\
UNINEXT-L-noObjPre && 61.8 & 85.4 & 68.1 & 55.0 & 68.1 \\
\system-L && 63.9 & 88.0 & 70.1 & 55.1 & 68.5 \\
\bottomrule
\end{tabular*}
\end{table}

\subsubsection{Video Instance Segmentation}
Video instance segmentation (VIS) shares a common objective with MOTS, yet extends its scope to encompass a broader array of object categories and diverse open scenarios. We evaluate \system on YTVIS 2019~\cite{yang2019video}, which contains 2,238 training, 302 validation, and 343 high-resolution video clips sourced from YouTube, encompassing a wide spectrum of real-world scenarios. Since most existing unified tracking models cannot support the VIS task, we only compare \system with the VIS-specific models and UNINEXT~\cite{yan2023universal}. 

In Table~\ref{tab:vis}, we present a detailed quantitative assessment, employing standard metrics including Average Precision (AP), AP at IoU 0.50 (AP50), AP at IoU 0.75 (AP75), Average Recall at IoU 1 (AR1), and Average Recall at IoU 10 (AR10). Note that we reproduce IDOL-L~\cite{wu2022defense} using the officially released code and report both the reproduced results (IDOL-L$^{*}$) and the results in their paper (\textcolor{gray}{IDOL-L}). Impressively, we can see that \system outperforms the existing methods, including UNINEXT-L-noObjPre, by a clear margin on all the metrics. These results show the remarkable capacity of \system to excel in the demanding domain of video instance segmentation, further substantiating its adaptability and proficiency across an extensive spectrum of tracking and segmentation tasks.

\subsection{Discussion and Analysis}
\subsubsection{Joint training \textit{v.s.} independent training} 
\system is jointly trained on various tracking datasets by alternatively feeding the batched data from different datasets during training. To verify the effects of joint training on task unification, we also train our model on different tasks separately to obtain several task-specific models. The comparison results on several representative datasets between joint training (Our-Joint), separate training (Ours-Sep), and Unicorn~\cite{yan2022towards}, are reported in Table~\ref{tab:joint}. We can see that joint training achieves consistently better results than both separate training and Unicorn on all the tasks. We hypothesize this is because we achieve a greater unification across different tasks, and training on the data from various sources could improve the generalization ability of our model. In addition, the comparison of Frames-Per-Second (FPS) also shows that \system enjoys a prominent advantage over Unicorn in terms of inference efficiency, \eg, 20.9 $vs$ 41.7 using Swin-Tiny as the backbone.
 
\begin{table}[!ht]
\centering
\caption{The performance on various tracking tasks with and without the proposed RFE module. TrNet: TrackingNet, D17: DAVIS 2017, YT19: YTVIS 2019. }
\label{tab:joint}
\renewcommand\arraystretch{1.1}
\setlength{\tabcolsep}{0.pt} % let TeX compute the intercolumn space
\begin{tabular*}{\linewidth}{@{\extracolsep{\fill}}lc | ccccccc @{}}
\toprule
\small
\multirow{2}*{\textbf{Method}} && \textbf{TrNet} & \textbf{D17} & \multicolumn{2}{c}{\textbf{MOT17}} & \textbf{YT19} & \multirow{2}*{FPS} \\
~ && P$_{norm}$ & $\mathcal{J\&F}$ & MOTA & IDF1 & mAP & ~ \\
\midrule
Unicorn-T~\cite{yan2022towards} && 82.0 & 62.7 & 71.7 & 73.4 & - & 20.9 \\
\midrule
Ours-Sep && 82.0 & 65.7 & 73.1 & 75.0 & 58.3 & 41.7 \\
Ours-Joint && \textbf{82.5} & \textbf{66.2} & \textbf{73.2} & \textbf{75.4} & \textbf{58.8} & 41.7 \\
\bottomrule
\end{tabular*}
\end{table}

\subsubsection{Impact of different components} 
In Table \ref{tab:ablation}, we conduct the ablation analysis of different components in our framework. We first discard the ReID loss (``w/o ReID"). We only perform the joint training due to the limited time. The results show that ReID loss can significantly improve the results of most tasks as it is the key to learning the association between objects. We also explore the effects of history size by only retaining the most recent query (denoted as ``w/o Mem") and Kalman filter (``w/o KF"). The performance degradation on the IT tasks is more remarkable for MR, implying a higher reliance on historical information. We also notice KF is very useful for MOT (which is widely adopted for previous methods including Unicorn). However, with it removed, the performance improvement of \system over Unicorn is even higher (Unicorn picks the box with the highest score without cross-frame association for IT, leading to differentiated inference pipelines for different tasks).

\begin{table}[!ht]
\centering
 \caption{The performance on various tracking tasks without ReID loss during training (w/o ReID), without Memory and Kalman filter during inference (w/o Mem and w/o KF). TrNet: TrackingNet, D17: DAVIS 2017, YT19: YTVIS 2019. }
\label{tab:ablation}
\renewcommand\arraystretch{1.1}
\setlength{\tabcolsep}{0.pt} % let TeX compute the intercolumn space
\begin{tabular*}{\linewidth}{@{\extracolsep{\fill}}lc | cccc @{}}
\toprule
\small
\multirow{2}*{\textbf{Method}} && \textbf{TrNet} & \textbf{D17} & {\textbf{MOT17}} & \textbf{YT19} \\
~ && P$_{norm}$ & $\mathcal{J\&F}$ & MOTA & mAP \\
\midrule
w/o ReID && 50.0 & 39.7 & 68.4 & 35.8 \\
w/o Mem && 79.3 & 62.8 & 72.0 & 58.1 \\
w/o KF && 80.1 & 64.2 & 59.7 & 57.3 \\
Unicorn-T w/o KF && - & - & 55.6 & -\\
Ours && \textbf{82.5} & \textbf{66.2} & \textbf{73.2} & \textbf{58.8} \\
\bottomrule
\end{tabular*}
 
%\vspace{-0.05in}
\end{table} 

\subsubsection{Impact of the RFE Module} 
RFE supplements the detector with appearance priors by modeling the correlation between target objects (or the reference frame) and the tracking frame with cross-attention. To evaluate its contribution, we remove it from \system and conduct experiments on various tasks. As shown in Table \ref{tab:rfe}, without the RFE module, the P$_{norm}$ on TrackingNet and the MOTA on MOT17 decrease by 1.6\% and 0.8\%, respectively. The performance degradation validates that the RFE module effectively improves the performance of our model.

\begin{table}[!ht]
\centering
\caption{The performance on various tracking tasks with and without the proposed RFE module. TrNet: TrackingNet, D17: DAVIS 2017, YT19: YTVIS 2019. }
\label{tab:rfe}
\renewcommand\arraystretch{1.1}
\setlength{\tabcolsep}{0.pt} % let TeX compute the intercolumn space
\begin{tabular*}{\linewidth}{@{\extracolsep{\fill}}lc | ccccc @{}}
\toprule
\small
\multirow{2}*{\textbf{Method}} && \textbf{TrNet} & \textbf{D17} & \multicolumn{2}{c}{\textbf{MOT17}} & \textbf{YT19} \\
~ && P$_{norm}$ & $\mathcal{J\&F}$ & MOTA & IDF1 & mAP \\
\midrule
w/o RFE && 80.9 & 64.7 & 72.4 & 74.6 & 58.5 \\
w/ RFE (Ours) && \textbf{82.5} & \textbf{66.2} & \textbf{73.2} & \textbf{75.4} & \textbf{58.8} \\
\bottomrule
\end{tabular*}
\vspace{-0.05in}
\end{table}

\begin{figure}[!h]
\centering
\includegraphics[width=\linewidth]{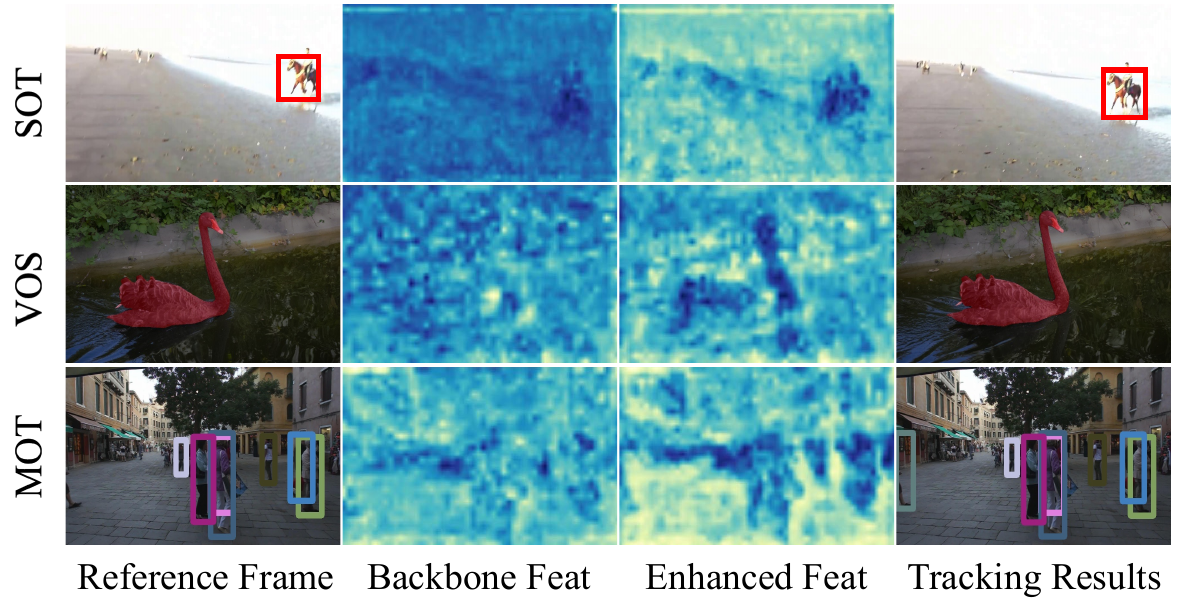}
\vspace{-0.2in}
\caption{Visualizations of the reference frame, backbone features, enhanced features, and the tracking results.}
\label{fig:vis_rfe}
\end{figure}

We also visualize the feature maps before and after the RFE module in Figure~\ref{fig:vis_rfe} by performing average pooling along the feature dimension. The results demonstrate that the target objects in the enhanced features are more distinguishable, with which our detector could localize the target objects more accurately.

We posit the reason why the RFE module brings more substantial improvements for the instance tracking (IT) tasks than category tracking (CT) tasks is the target object in IT is specified by the reference frame and can belong to any category, while the target objects for existing CT setting belong to specific close-set categories. So the IT performance relies heavily on the trajectory and the CT performance relies more on the detector. To further illustrate this, we evaluated \system on videos belonging to the COCO category (\eg, horse) and not (\eg, camel) on DAVIS17. The $\mathcal{J\&F}$ values are reported in Table \ref{tab:coco}.

\begin{table}[!ht]
\centering
\caption{The performance on the videos belonging to the COCO category (\eg, horse) and not (\eg, camel) on DAVIS17. }
\label{tab:coco}
\renewcommand\arraystretch{1.1}
\setlength{\tabcolsep}{0.pt} % let TeX compute the intercolumn space
\begin{tabular*}{\linewidth}{@{\extracolsep{\fill}} lc | ccc @{}}
\toprule
\textbf{Method} && COCO Categories & Others & Overall \\
\midrule
w/o RFE && 69.7 & 59.7 & 64.7 \\
w/ RFE && 70.5 ($\uparrow$ 0.8) & 61.9 ($\uparrow$ 2.2) & 66.2 ($\uparrow$ 1.5) \\
\bottomrule
\end{tabular*}
\end{table}

We can see that RFE has a more prominent influence on the latter. Therefore, we believe it can bring more significant performance gains for CT tasks in the open-vocabulary setting, which is more challenging and of higher practical value.

\subsubsection{Impact of backbone networks} 
We adopt the Swin transformer as the backbone since it is a popular choice for previous tracking models, \eg, SwinTrack~\cite{lin2021swintrack}, IDOL~\cite{wu2022defense}, SeqFormer~\cite{wu2022seqformer}, \etal. In Table \ref{tab:backbone}, we also employed ConvNext~\cite{liu2022convnet} Large as the backbone for a fair comparison with Unicorn~\cite{yan2022towards} and UNINEXT~\cite{yan2023universal}. The results demonstrate that overall, using ConvNext-L will lead to better results compared to Swin-L, and \system still outperforms Unicorn and UNINEXT on all the tasks.

\begin{figure*}[t]
\centering
\includegraphics[width=\linewidth]{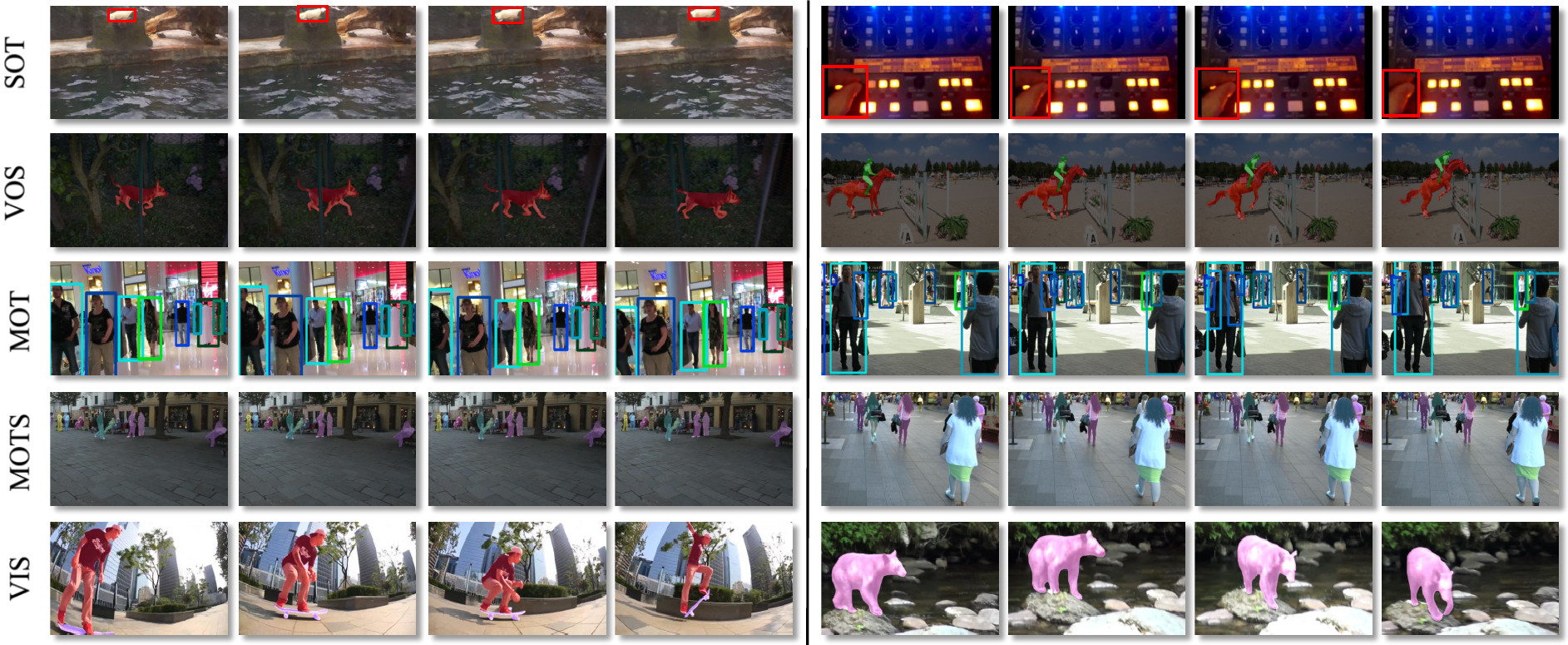}
\vspace{-0.25in}
\caption{Visualizations of tracking results predicted by \system on various tracking tasks. }
\label{fig:visualization}
\end{figure*}

\begin{table}[!ht]
\centering
\caption{Comparison between Unicorn and \system with different backbone networks.}
\label{tab:backbone}
\renewcommand\arraystretch{1.1}
\setlength{\tabcolsep}{0.pt} % let TeX compute the intercolumn space
\begin{tabular*}{\linewidth}{@{\extracolsep{\fill}}lc | cccc @{}}
\toprule
\small
\multirow{2}*{\textbf{Method}} && \textbf{TrNet} & \textbf{D17} & {\textbf{MOT17}} & \textbf{YT19} \\
~ && P$_{norm}$ & $\mathcal{J\&F}$ & MOTA & mAP \\
\midrule
Unicorn-ConvL && 86.4 & 69.2 & 77.2 & - \\
%UNINEXT-ConvL && 88.2 & 77.2 & - & 64.3 \\
UNINEXT-ConvL-noObjPre && 86.6 & 73.2 & - & 61.8 \\
Ours-SwinL && 86.7 & 71.0 & 79.1 & 63.9 \\
Ours-ConvL && \textbf{86.9} & \textbf{71.0} & \textbf{79.3} & \textbf{64.1}  \\
\bottomrule
\end{tabular*}
\vspace{-0.3in}
\end{table}

\subsubsection{Impact of hyperparameters}
We try different $\lambda_{1}$ and $\lambda_{2}$ to see their effects on the performance. As can be seen in Table~\ref{tab:hyper}, changing $\lambda_{1}$ and $\lambda_{2}$ will not significantly influence the results on various tasks, validating the effectiveness and robustness of the proposed method.

\begin{figure}[!ht]
\centering
\includegraphics[width=0.9\linewidth]{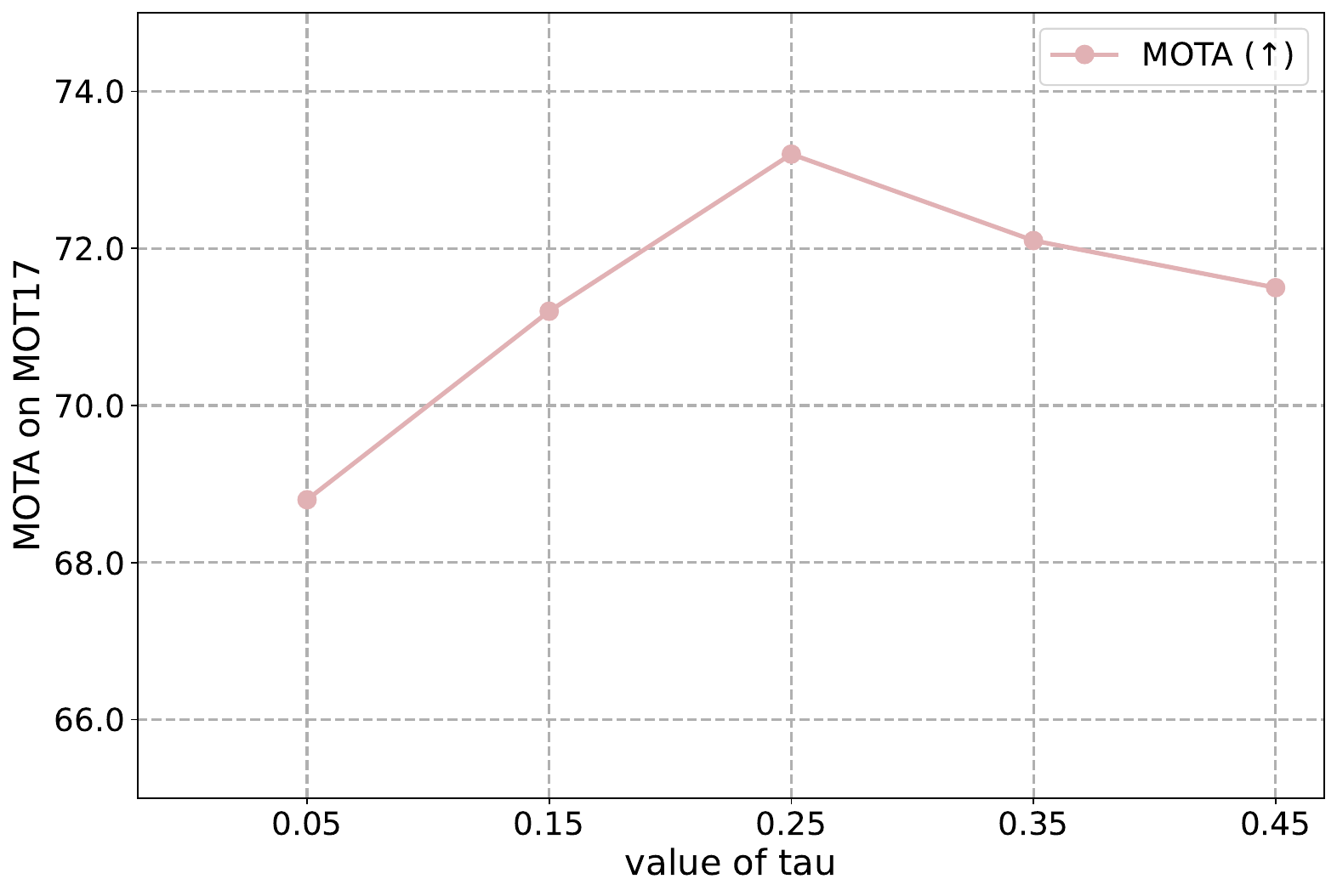}
\caption{MOTA on MOT17 test set with different $\tau$.}
\label{fig:tau}
\end{figure}

$\tau$ only impacts the inference of multiple object tracking tasks (\ie, MOT(S) and VIS). To show its effect, we evaluate our model on MOT17 test set with different values. Figure~\ref{fig:tau} shows that the choice of $\tau$ has a remarkable effect on MOT results, with 0.25 leading to the highest MOTA.

\begin{table}[!ht]
\centering
\caption{\system with different hyperparameters.}
\label{tab:hyper}
\renewcommand\arraystretch{1.1}
\setlength{\tabcolsep}{0.pt} % let TeX compute the intercolumn space
\begin{tabular*}{\linewidth}{@{\extracolsep{\fill}}lllc | cccccc @{}}
\toprule
\small
\multirow{2}*{\textbf{ $\lambda_{1}$ }} & \multirow{2}*{\textbf{ $\lambda_{1}$ }} & \multirow{2}*{\textbf{ $\beta$ }} && \textbf{TrNet} & \textbf{D17} & \multicolumn{2}{c}{\textbf{MOT17}} & \textbf{YT19}  \\
~ & ~ & ~ && P$_{norm}$ & $\mathcal{J\&F}$ & MOTA & IDF1 & mAP  \\
\midrule
2.0 & 2.0 & 10. && 82.5 & 66.2 & 73.2 & 75.4 & 58.8 \\
1.0 & 1.0 & 10. && 80.5 & 66.2 & 74.9 & 76.2 & 57.9 \\
3.0 & 3.0 & 10. && 81.7 & 65.6 & 75.2 & 77.2 & 58.1 \\
\bottomrule
\end{tabular*}
\end{table}

\subsection{The possibility to extend \system}
In recent years, some foundation models have been proposed in the field of vision (\eg, SAM~\cite{kirillov2023segment,ravi2024sam}) and multimodal domain (\eg, LLaVA~\cite{liu2023visual}). This section will discuss the possibility of integrating them into \system to extend its capabilities and functionality.

SAM2~\cite{ravi2024sam} is a foundation tracking model dedicated to solving all tracking tasks with one model, sharing the same goal as \system. While it is difficult to incorporate SAM2 into our framework directly, it is quite promising for us to borrow from their design, \eg, introducing dense memory to improve the performance of VOS, or using the large-scale, high-quality datasets that they have constructed to enhance the capabilities of our model. Large Multimodal Models (LMMs) can interpret and process diverse modalities in one framework. While \system focuses on object tracking by unifying tracking-with-detection, combining it with LMMs could open up exciting possibilities: 1) leveraging text queries to guide tracking or incorporating contextual cues from multimodal inputs to improve the robustness of \system in ambiguous scenarios. and 2) enabling advanced applications like generating natural language descriptions for tracklets.

While these integrations may lie outside the scope of the current work, they represent promising directions for future research to enhance tracking performance in complex multimodal environments further.

\begin{figure*}[!ht]
\centering
\includegraphics[width=\linewidth]{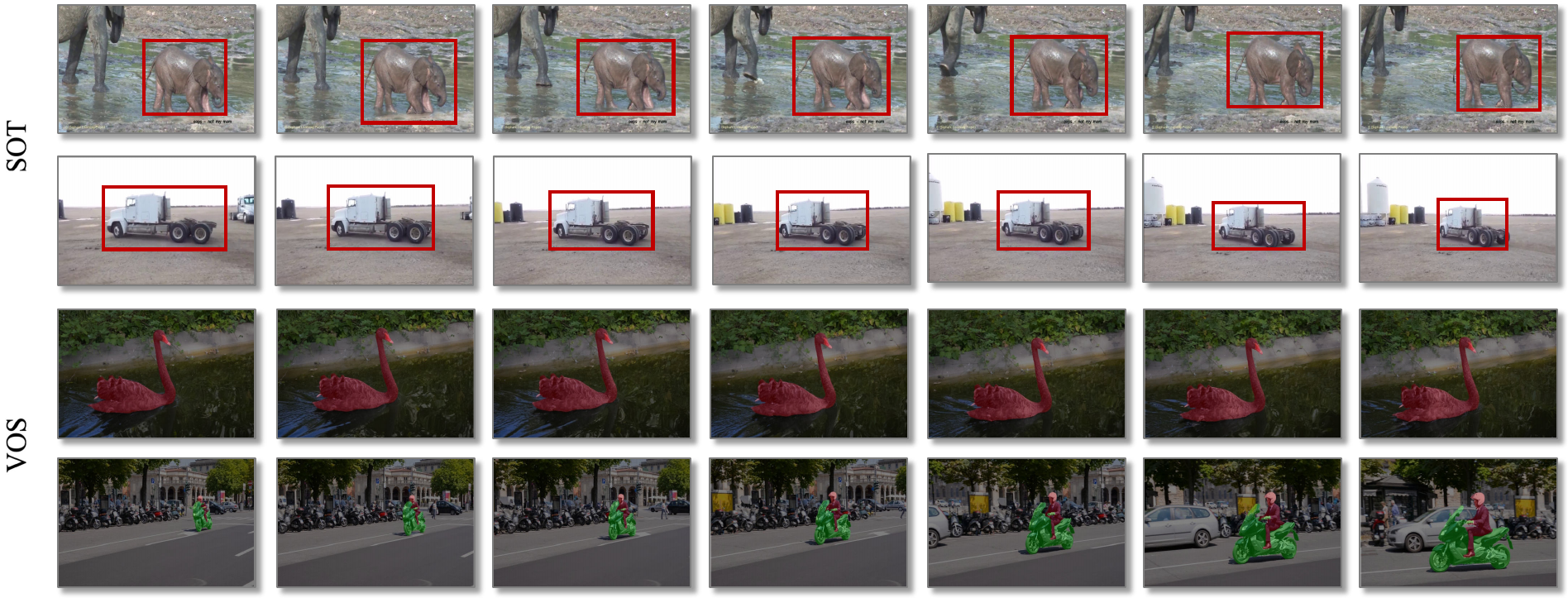}
\vspace{-0.25in}
\caption{Visualizations of tracking results predicted by \system on long videos. }
\label{fig:visualization_long}
\end{figure*}

\subsubsection{Visualizations} 
We visualize the tracking results of \system on different tasks in Figure~\ref{fig:visualization}. For the instance tracking tasks, the target objects may belong to any classes, \eg the luminous corner of a stone in the 1st row, which requires the model to possess strong detection and association ability. For the category tracking tasks, the dramatic movements of target objects  (5th row) and the severe occlusions (the 3rd row) both pose great challenges to the robustness of the model. The superior performance of \system on both types of tasks fully proves its effectiveness. Besides, we show the results of \system on long videos in Figure~\ref{fig:visualization_long}, from which we can see that our method can keep tracking the target object over a long time, no matter whether it is moving slowly (line 1 and 3) or rapidly (line 2 and 4).

\begin{figure}[!ht]
  \centering
   \includegraphics[width=\linewidth]{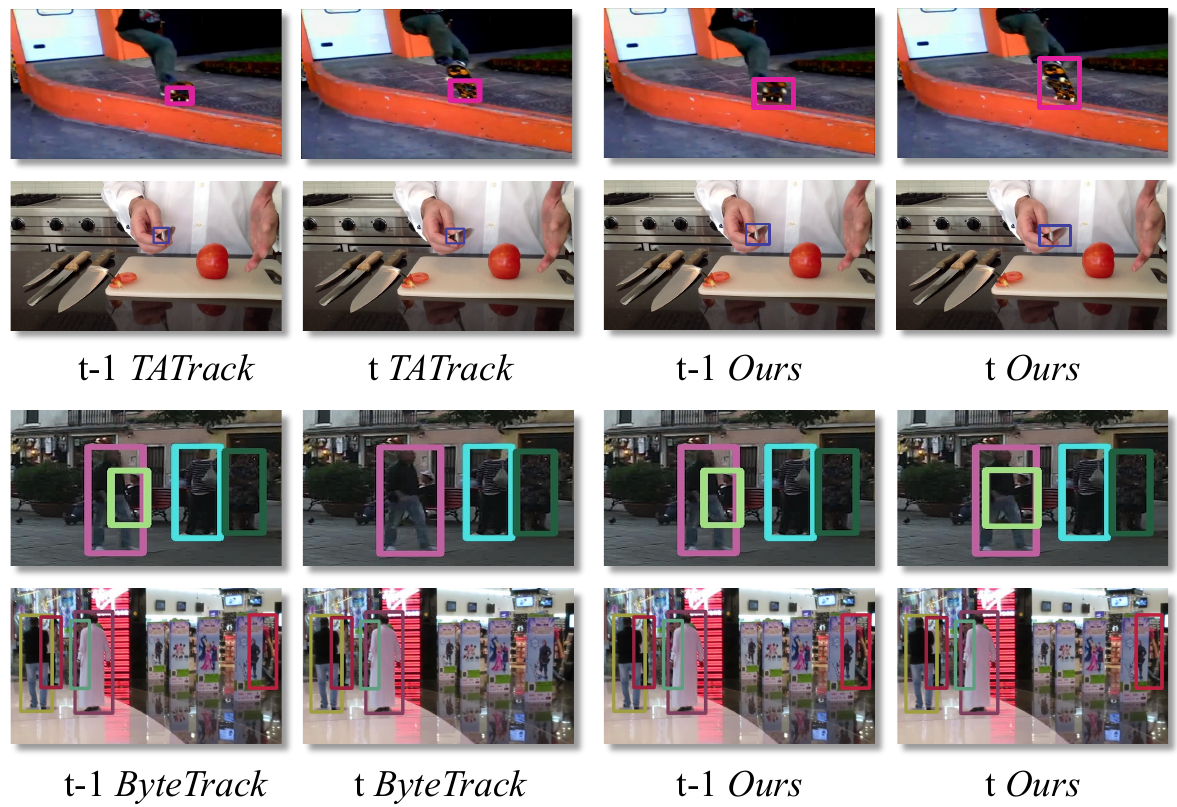}
   %\vspace{-0.25in}
   \caption{Comparison of the tracking results between \system (tracking-with-detection) and ByteTrack~\cite{zhang2022bytetrack} (tracking-by-detection), as well as TATrack~\cite{he2023target} (tracking-as-detection). }
   \label{fig:compare}
\end{figure}

In addition, to demonstrate the merits of our proposed tracking-with-detection paradigm in comparison to the existing tracking-as-detection and tracking-by-detection methodologies, we present a comparative analysis of tracking results produced by our method against two state-of-the-art task-specific models in Figure \ref{fig:compare}: TATrack~\cite{he2023target}, an advanced Single Object Tracking (SOT) model employing tracking-as-detection, and ByteTrack~\cite{zhang2022bytetrack}, a superior Multiple Object Tracking (MOT) model following tracking-by-detection. We can observe that \system not only detects the target object more accurately but also showcases enhanced robustness to occlusion. This can be credited to our tracking-with-detection paradigm, which detects the target object on the entire image and adeptly makes use of historical information, thereby leading to improved overall performance.

\subsubsection{Failure Case} 
\begin{figure}[!ht]
\centering
\includegraphics[width=\linewidth]{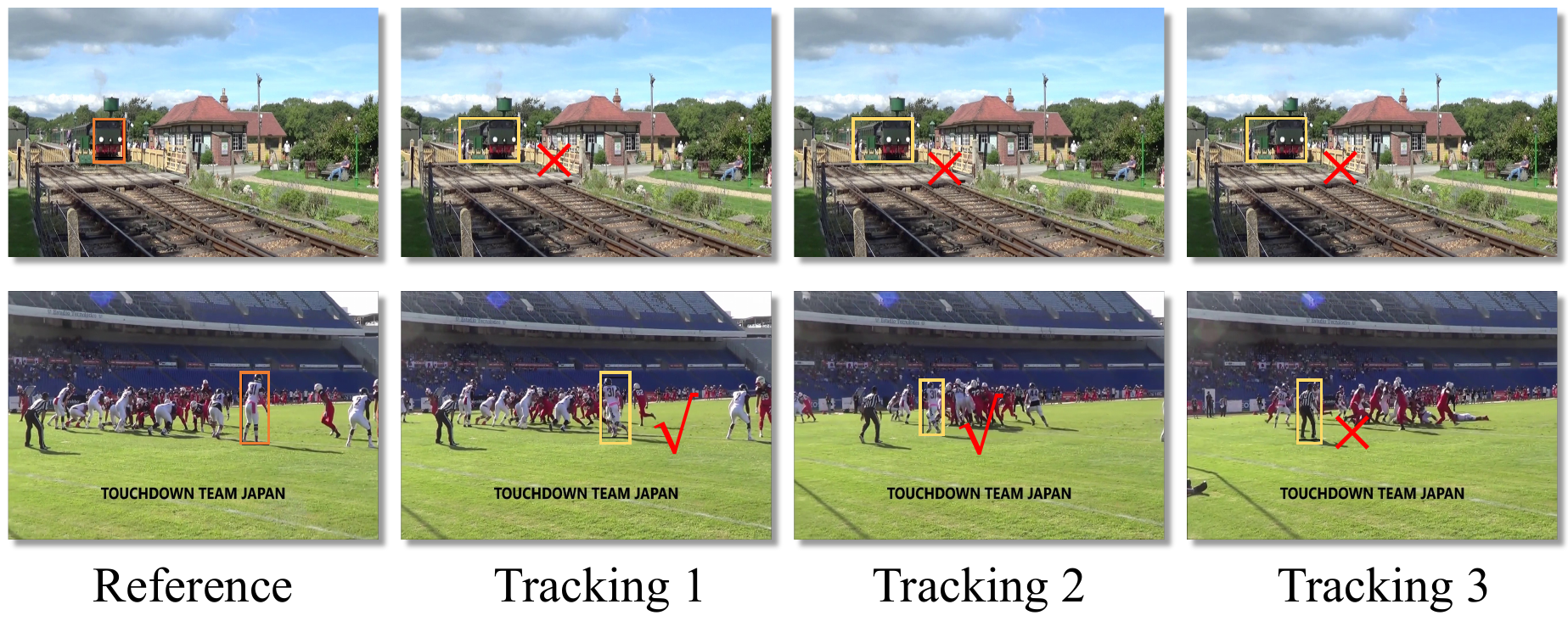}
\caption{Failure cases: boxes in orange are user-provided annotations, while boxes in yellow are predicted by \system.}
\label{fig:failure}
\end{figure}

Although \system demonstrates notable strengths on various tracking tasks, it may struggle in certain cases, offering insights into its behavior under specific conditions. We showcase some failure cases on TrackingNet in Figure~\ref{fig:failure} for analysis. 

First of all, our approach tends to track complete objects when the user annotates only part of an object rather than all of it in the first frame. This may be because we co-trained the model on both the instance tracking and category tracking datasets, thus giving it a preference for objects with complete semantics. Another failure situation occurs when, for example, the target object moves unusually fast or is occluded by other objects. In these cases, our approach encounters challenges in maintaining tracking continuity. This can be attributed to the degradation of the quality of the instance embedding, which is particularly noticeable in high-speed motion and occlusion cases. 

%-------------------------------------------------------------------------

\section{Conclusion}
\label{sec:conclusion}
This paper presented \system, a unified Deformable DETR-based tracking model that addresses both instance tracking tasks (\ie, SOT and VOS) and category tracking tasks (\ie, MOT, MOTS, and VIS), with a fully shared network architecture, model weights, and inference pipeline. Incorporating the advantages of the dominant solutions in the above two types of tasks, we further introduced a \textbf{tracking-with-detection} paradigm, where tracking supplements detection with appearance priors to locate the targets more accurately and detection provides tracking with candidate boxes for the association. Extensive experiments conducted on a wide variety of tracking benchmarks demonstrate the effectiveness of the proposed method. 

Although \system significantly outperforms Unicorn~\cite{yan2022towards} on the VOS task, there is still a gap between the task-specific models. We believe the reason lies in that they widely adopt high-resolution spatial-temporal memory for dense matching, whereas we only exploit compact query-based memory. In the future, we will explore the combination of both types of memory for tracking to further improve the performance on the VOS task.

\noindent \textbf{Acknowledgement.} This work was supported in part by National Natural Science Foundation of China (\#62032006).

\bibliographystyle{IEEEtran}
\bibliography{IEEEfull}

\end{document}